\documentclass[preprint,5p,times,twocolumn,authoryear]{elsarticle}

\usepackage{amssymb}
\usepackage[caption=false,font=footnotesize,labelfont=footnotesize,textfont=footnotesize,subrefformat=parens]{subfig}
\usepackage{booktabs}
\usepackage{color}
\usepackage{hyperref}
\usepackage[hang]{footmisc}

\journal{ISPRS Journal of Photogrammetry and Remote Sensing}

\begin{document}

\begin{frontmatter}



\title{SuperMapNet for Long-Range and High-Accuracy Vectorized HD Map Construction}


\author[inst1]{Ruqin Zhou}
\author[inst1]{Chenguang Dai}
\author[inst2]{Wanshou Jiang}
\author[inst1]{Yongsheng Zhang}
\author[inst1]{Hanyun Wang}
\author[inst3]{San Jiang \corref{cor1}}

\cortext[cor1]{
Corresponding author at: Guangdong Key Laboratory of Urban Informatics, Shenzhen University, Shenzhen, 518060, Guangdong, China.\\
\textit{E-mail address}: jiangsan@szu.edu.cn (S. Jiang)
}

\affiliation[inst1]{organization={School of Surveying and Mapping, Information Engineering University},
            city={Zhengzhou},
            postcode={450001}, 
            state={Henan},
            country={China}}

\affiliation[inst2]{organization={State Key Laboratory of Information Engineering in Surveying, Mapping and Remote Sensing, Wuhan University},
            city={Wuhan},
            postcode={430079}, 
            state={Hubei},
            country={China}}
            
\affiliation[inst3]{organization={Guangdong Key Laboratory of Urban Informatics, Shenzhen University},
            city={Shenzhen},
            postcode={518060}, 
            state={Guangdong},
            country={China}}

\begin{abstract}
Vectorized HD map construction is formulated as the classification and localization of typical map elements according to features in a BEV space, which is essential for autonomous driving systems, providing interpretable environmental structured representations for decision and planning. Remarkable work has been achieved in recent years, but there are still major issues: (1) in the generation of the BEV features, single modality-based methods suffer from limited perception capability and range, while direct concatenation-based multi-modal fusion inadequately exploits cross-modal synergies and fails to resolve spatial disparities, resulting in incomplete BEV representations with feature holes; (2) in the classification and localization of map elements, existing methods overly rely on point-level modeling information while neglecting the interaction between elements and interaction between point and element, leading to erroneous shapes and element entanglement with low accuracy. To address these limitations, we propose SuperMapNet, a multi-modal framework for long-range and high-accuracy vectorized HD map construction. It uses both camera images and LiDAR point clouds as input, and first tightly couple semantic information from camera images and geometric information from LiDAR point clouds by a cross-attention based synergy enhancement module and a flow-based disparity alignment module for long-range BEV feature generation. Subsequently, local features from point queries and global features from element queries are tightly coupled by three-level interactions for high-accuracy classification and localization, where Point2Point interaction captures local geometric consistency between points of the same element, Element2Element interaction models global semantic relationships between elements, and Point2Element interaction complement element information for its constituent points. Experiments on the nuScenes and Argoverse2 datasets demonstrate superior performances, surpassing SOTAs over 14.9/8.8 mAP and 18.5/3.1 mAP under the hard/easy settings, respectively. The code is made publicly available \tnoteref{t1}.
\end{abstract}

\tnotetext[t1]{\url{https://github.com/zhouruqin/SuperMapNet}}




\end{frontmatter}


\section{Introduction}

As the core component of environmental perception in autonomous driving systems, the high-definition (HD) map provides an interpretable structured representation of neighbor environment for decision and planning by integrating centimeter geometric topology and semantic traffic attribute of map elements, such as road boundaries, lane dividers, and pedestrian crossings \citep{machmap, streammapnet}. Consequently, constructing HD map with high accuracy and long range is critical to ensuring the intelligence, reliability, and safety of autonomous driving systems in complex urban scenarios \citep{mapex, LUO2023122, GevBEV}.

\begin{figure}[t]
	\centering
	\subfloat[rasterized map predicted by SuperFusion \citep{superfusion}, which assigns each pixel a label.]{\includegraphics[width=0.45\textwidth, height = 70pt]{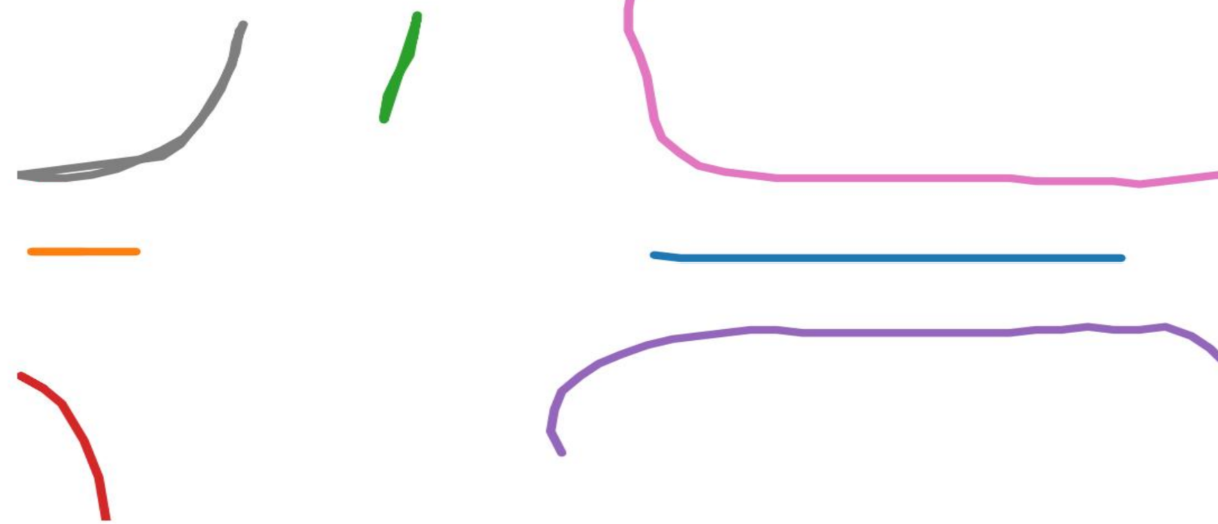}%
		\label{a}}
	\hfil
	\subfloat[vectorized map predicted by PivotNet \citep{pivotnet}, which models map elements in the form of ordered keypoint sets.]{\includegraphics[width=0.45\textwidth, height = 70pt]{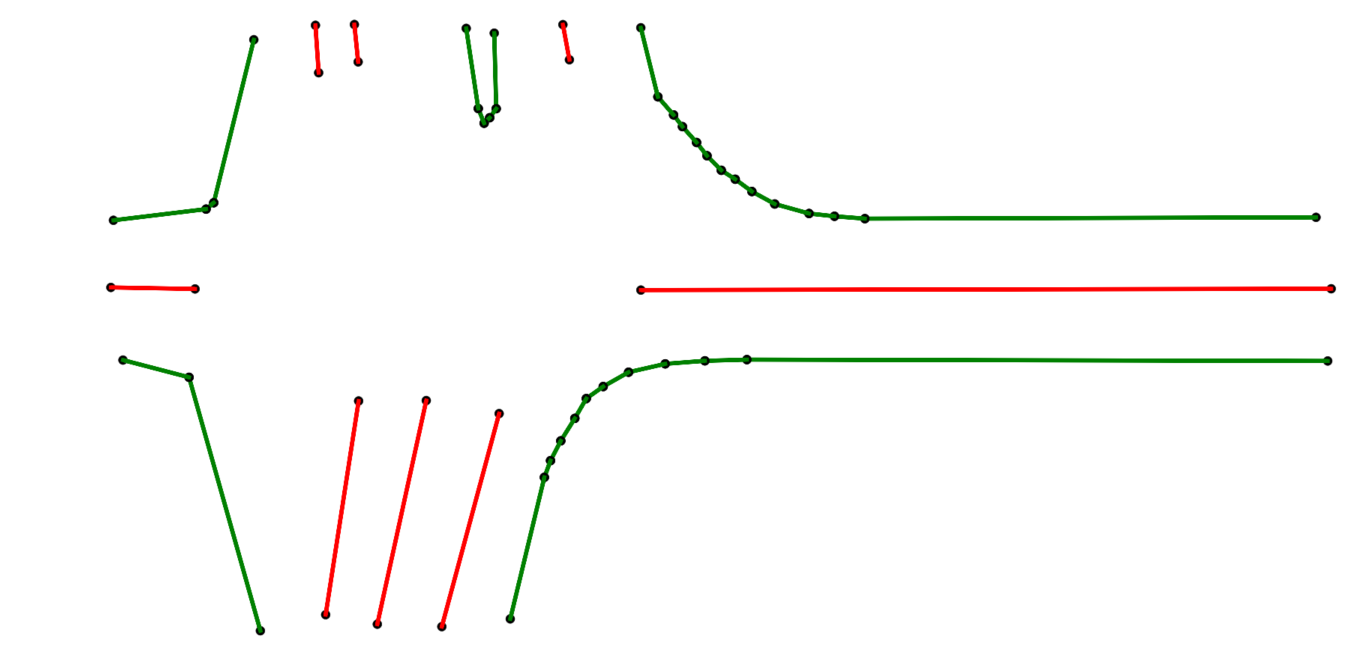}%
		\label{b}}
	\caption{Visualization comparison between rasterized and vectorized HD map.}
	\label{fig:raster}
\end{figure}

Traditional methods rely on an off-line map construction strategy, which first employ SLAM (Simultaneous Localization and Mapping) algorithms to generate a large-scale global point cloud \citep{slam}, followed by manual annotation to create globally consistent semantic maps \citep{hdmapnet}. However, these approaches faces significant challenges in terms of high cost and low temporal freshness \citep{vectormapnet}, making them difficult to adapt to dynamic urban scenarios. Recent advancements in hardware and data processing algorithms have enabled onboard sensor-based local HD map construction as a promising alternative. HDMapNet \citep{hdmapnet}, the first deep learning-based framework for HD map construction, formulates the HD map construction as a semantic segmentation task by rasterizing the HD map and assigning each pixel a label. By decoding features in the BEV space, three sub-tasks are simultaneously implemented by the framework: semantic segmentation, instance embedding, and direction estimation. HDMapNet \citep{hdmapnet} has established the foundational architecture for subsequent raster-based map segmentation methods \citep{maptrv2, superfusion, bevfusion}. \citet{diffmap} integrated a diffusion model into the basic framework and proposed DiffMap, which learns map priors by iteratively adding and removing noise to align outputs to real-time observations. However, rasterized map is not an ideal representation. It inherently contains redundant information of each pixel,demanding substantial storage resources, especially when the map's range is large \citep{pivotnet}. Furthermore, its assumption of pixel-level independence neglects structural relationships between pixels and elements \citep{vectormapnet}, often resulting in incomplete or distorted shapes (as shown in Fig. \ref{a}). Additionally, complex post-processing for vectorization is required for subsequent tasks, introducing computational overhead and error accumulation for downstream tasks.

To overcome the aforementioned limitations caused by the rasterized representation, constructing vectorized HD maps in an end-to-end manner has recently emerged as an increasingly prominent solution with notable success. Early approaches  \citep{maptr, maptrv2, admap, gemap} have attempted to model map elements in the form of uniform ordered point sets, and learn the label and location of each point directly by models. Nevertheless, this uniform point modeling strategy struggles to balance computational efficiency and construction accuracy \citep{pivotnet}. In response, a novel solution \citep{vectormapnet, pivotnet, bemapnet} that models map elements as ordered keypoints, as shown in Fig. \ref{b}, has been proposed in the last two years, greatly improving storage efficiency. However, current keypoint-based methods only use single modality with limited perception capability, or directly concatenate features of different modalities without considering synergies and disparities for the generation of BEV features, restricting effective perception range (usually 60 $m$ in the Y-axis). In addition, only point information is utilized for classification and localization of map elements, which struggles to handle element failures, such as erroneous element shapes or entanglement between elements, leading to low construction accuracy.

\begin{figure*}[]
    \centering
    \includegraphics[width=\linewidth, ]{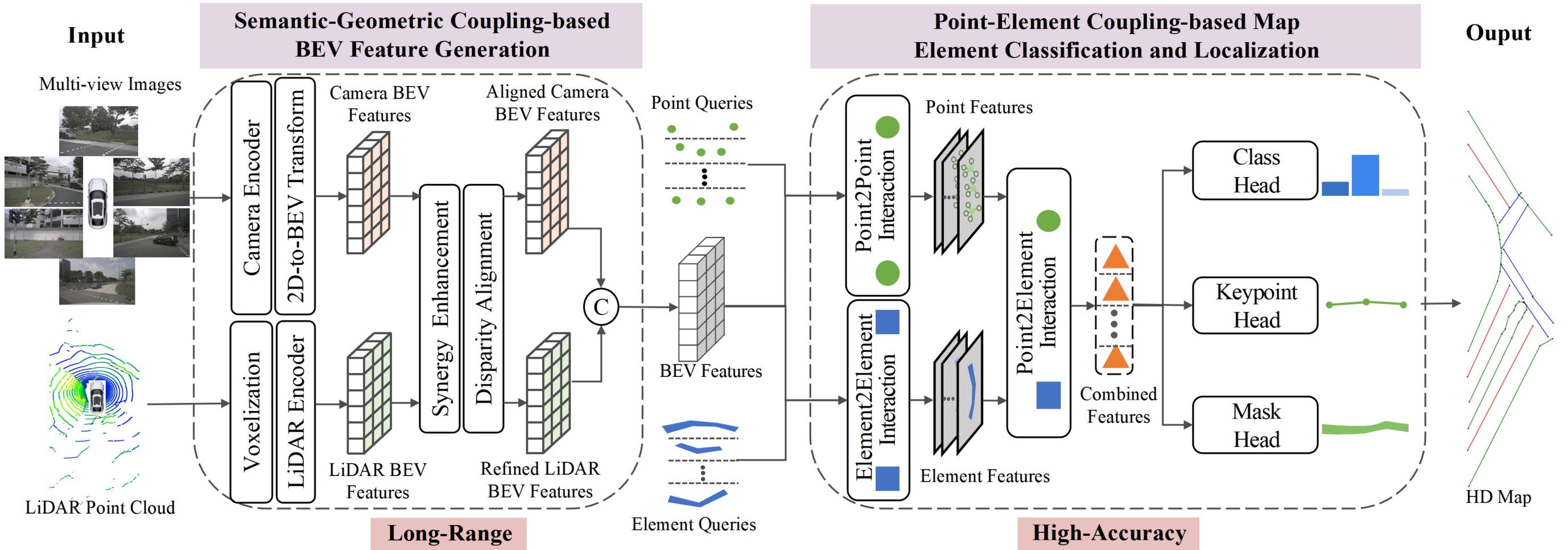}
    \caption{The overall architecture of SuperMapNet. Semantic information from camera images and geometric information from LiDAR point clouds are tightly couple by a cross-attention based synergy enhancement module and a flow-based disparity alignment module for long-range BEV feature generation, and local features from point queries and global features from element queries are tightly coupled by three-level interactions, Point2Point, Element2Elment, and Point2Element, for high-accuracy classification and localization.}
    \label{fig:framework}
\end{figure*}

In this study, we introduce SuperMapNet, a multi-modal framework for long-range and high-accuracy vectorized HD map construction. The overall architecture is illustrated in Fig. \ref{fig:framework}. Both camera images and LiDAR point clouds are used as input, and through the feature encoder of each modality, camera BEV features with semantic information and LiDAR BEV features with geometric information are obtained, respectively. Camera BEV features and LiDAR BEV features with geometric information are first tightly coupled by a cross-attention based synergy enhancement module and a flow-based disparity alignment module to learn fused BEV features with rich semantic and geometric information in a long range. Secondly, local knowledge from point queries and global knowledge from element queries are tightly coupled by interactions at three levels for high-accuracy classification and localization of map elements, where Point2Point interaction for local geometric consistency learning between points of the same element and of each point, Element2Element interaction for semantic constraints learning between different elements and of each elements, and Point2Element interaction bridges local point details with global element context to resolve ambiguities. Our contributions are summarized as follows:

\begin{itemize}
\item {\textbf{Long-Range:} with the consideration of synergies and disparities between camera images and LiDAR point clouds, SuperMapNet maintains remarkable performance over long ranges, up to 120 $m$ in the Y-axis, which is twice the perception range of other comparative methods;}
\item {\textbf{High-Accuracy:} interactions at three levels between point queries and element queries effectively reduce erroneous shapes and entanglement between elements, outperforming previous SOTAs by 14.9/8.8 mAP and 18.5/3.1 mAP on nuScenes and Argoverse2 datasets under hard/easy settings, respectively.}
\end{itemize}

This paper is organized as follows. Section 2 reviews the related work for HD map construction. Section 3 presents the workflow of SuperMapNet, and the proposed semantic and geometric information coupling module and point and element information coupling module. The evaluation and analysis on nuScenes and Argoverse2 datasets of the proposed SuperMapNet are conducted in Section 4, with the comparison to SOTAs. Section 5 discusses the ablation study of different modules. Finally, Section 6 presents the conclusions.

\section{Related Work}
The main workflow of HD map construction can be divided into two steps: generation of BEV features and classification and localization of map elements. The former aims to convert features from various views and various modalities into a unified BEV space, and the latter aims to infer the shape, location, and category of each map element based on BEV features.

\subsection{Multi-modal Fusion} 
\begin{figure*}[tp]
    \centering
    \includegraphics[width=\linewidth]{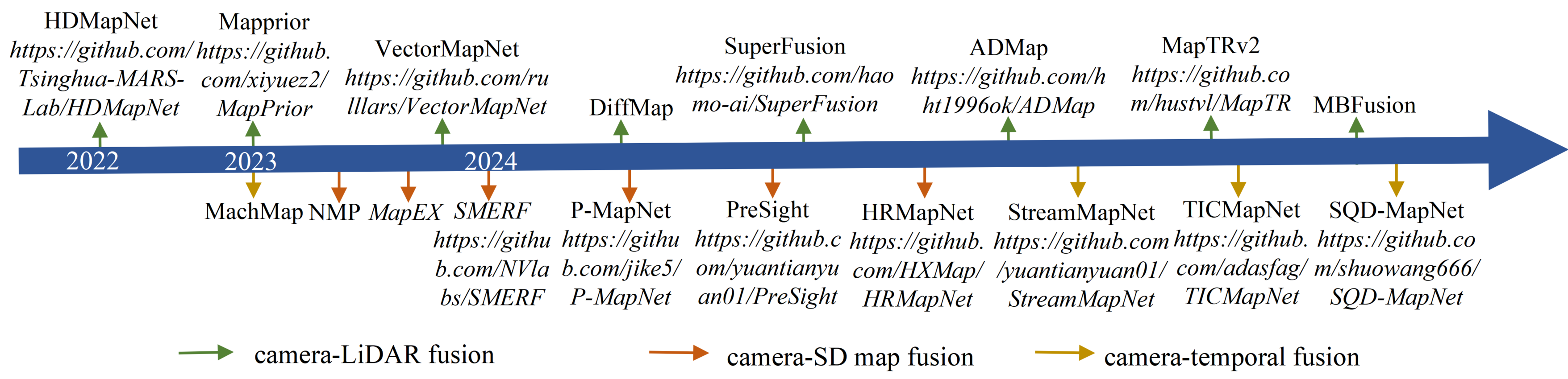}
    \caption{The evolution of deep learning-based  HD map construction methods with multi-modalities. According to different modalities used for the generation of BEV features, the existing methods are divided into three types: camera-LiDAR fusion, camera-SD Map fusion, and camera-temporal fusion methods.}
    \label{fig:multi-modal}
\end{figure*}

According to different modalities used for the generation of BEV features, as shown in Fig. \ref{fig:multi-modal}, the existing local HD map construction methods can be divided into three types: camera-LiDAR fusion, camera-standard definition map (SD Map) fusion, and camera-temporal fusion methods.

\textbf{Camera-LiDAR Fusion}: HDMapNet \citep{hdmapnet} pioneered the first camera-LiDAR fusion framework for HD map construction, aiming to make up for the perception limitations of single modality. This framework first employs  geometric projection\citep{bev}, such as IPM \citep{ipm} and LSS \citep{lss}, or Transformer-based methods, such as BEVFormer \citep{bevformer}, TransFuser \citep{transfusion}, PolarFormer \citep{pollarformer}, and WidthFormer \citep{widthformer}, to explicitly or implicitly transform features in perspective views into a bird's-eye view with the prior geometric information of the visual camera. At the same time, models, such as PointPillar \citep{pointpillar}, Second \citep{sec} or VoxelNet \citep{VoxelNet}, are used to extract BEV features from LiDAR point clouds after voxelization; then, the LiDAR BEV features and camera BEV features are directly concatenated to generate fused features. However, this direct concatenation strategy overlooks synergies and disparities between different modalities, resulting in limited ranges and low accuracy. SuperFusion \citep{superfusion} explicitly leverages the geometric information of LiDAR point clouds to supervise depth estimation of camera images, and uses a cross-attention mechanism to integrate the geometric information of point clouds with semantic information of camera images, thereby alleviating the problem of insufficient details in the distance caused by sparse point clouds. MBFusion \citep{mbfusion} adds a dual dynamic fusion module based on the cross-attention mechanism to automatically select valuable information from different modalities for better feature fusion. Camera-LiDAR fusion methods can significantly improve the accuracy and robustness of the model by complementing geometric and semantic information, but existing methods ignore the feature misalignment caused by positioning errors of different sensors.

\textbf{Camera-SD Map Fusion}: NMP \citep{nmp} presents a novel neural prior network-based paradigm for HD map construction, aiming to enhance lane perception and topological understanding by leveraging prior knowledge including explicit standard-definition (SD) map data and implicit temporal cues, thereby improving perception capabilities under adverse weather conditions and at extended ranges. The framework first queries the current local SDMap through the ego-pose of vehicles, and then encodes local SD map prior knowledge and online sensor perception information through models such as CNN and Transformer. By calculating the cross-attention between local SD map prior knowledge and BEV features of online sensors, local HD map is constructed. NMP \citep{nmp} has served as a foundational inspiration for subsequent researches 
\citep{p-mapnet, mapex, SMERF, hrmapnet}. On this basis, PreSight \citep{PreSight} constructs a city-level neural radiation field and incorporates foundational vision model DINO \citep{dino} to embed generalizable semantic priors, and then online BEV features with semantic priors are directly concatenated to enhance perception ability. The camera-SD Map fusion methods effectively enhance the accuracy, robustness, and perceptual range through the complementarity of static and dynamic information, geometric and semantic information, significantly reducing real-time computational burden. However, this strategy relies on the availability of prior maps, and requires pre-storage of large-scale city-level SD Map data, resulting in high storage costs. 

\textbf{Camera-temporal fusion method}: MachMap \citep{machmap} first proposed a temporal fusion strategy, which uses ego-pose of the vehicle to associate the previous hidden states of BEV features with the current hidden state, and then obtains the fused features through concatenation. This strategy aims to utilize temporal information of continuous frames to solve the problem of data loss caused by occlusion and other complex scenes. On this basis, StreamMapNet \citep{streammapnet} and StreamMapNet SQD \citep{sqd} proposed a streaming temporal fusion strategy, which encodes all historical information into memory features to save costs and establish long-term temporal correlations, and then uses gated recurrent units (GRU) to fuse BEV features at different time points. Considering the differences in feature coordinates at different times, TICMapNet \citep{TICMapNet} designed a temporal feature alignment module to eliminate coordinate errors, and then fused temporal information through a deformable attention mechanism. The camera-temporal fusion method fills feature holes and expands perception range through dynamic static complementarity, but research on this strategy is still in its infancy.

\subsection{Vectorized HD Map Construction} 
 Vectorized HD map construction models map elements as ordered point or line sets, querying modeling information based on  BEV features for localization and classification of map elements. According to their core innovations, existing vectorized HD map construction methods can be divided into two categories: innovations in the representation of map elements and innovations in the modeling information.

\textbf{Innovations in the representation of map elements}: MapTR \citep{maptr} is the first end-to-end vectorized map construction framework, modeling map elements using uniform ordered point sets. It encodes element features through hierarchical query embeddings, and performs hierarchical matching through permutation-equivariant modeling. It established the theoretical foundation for the representation of vectorized map element based on uniform ordered point sets, which has been systematically adopted in subsequent approaches \citep{maptrv2, admap, gemap}. However, this uniform point set modeling strategy struggles to balance computational complexity and construction accuracy \citep{pivotnet}. To tackle this problem, \cite{pivotnet} proposed a new framework, named PivotNet, and introduced a unified representation based on ordered keypoint sets for map element modeling. Uniform ordered point sets are split into two sequences, where pivot sequence contains points that critical for preserving element shape and direction, and collinear sequence refer to points that can be pruned without affecting geometry. The keypoint-based method can optimize storage efficiency and maintain the construction accuracy. VectorMapNet \citep{vectormapnet} proposed a coarse-to-fine construction strategy, which models map elements in the form of curves. It first predicts keypoints of map elements, and refines them into polylines through Transformer architecture. BeMapNet \citep{bemapnet} models map elements as piecewise B\'ezier curves, and designs a piecewise B\'ezier head for dynamic curve modeling with two branches for classification and regression, where the first branch predicts the number of curve piece to determine overall length, and the second estimates the coordinates of control points to determine the curve shape. However, above methods rely solely on point information for the classification and localization of map elements, making it difficult to handle element failures, such as erroneous shapes or entanglement between elements \citep{himapnet}.

\textbf{Innovations in the modeling information}: To address the limitations of only using point-level modeling information, MapTRv2 \citep{maptrv2} introduces hybrid queries by integrating element queries with point queries. Self-attention mechanisms are applied along point dimension and element dimension to extract fine-grained geometric features and semantic relationships, respectively. Experiments demonstrates the great potential of element information in dealing with element failures. Similarly, InsMapper \citep {insMapper} combines element queries with point queries through a hybrid query generation scheme, and treats hybrid queries as the basic processing units for the query of modeling information. However, this results in computational redundancy, where element queries are redundantly reused for point-level tasks (and vice versa), significantly increasing computational complexity. On the basis, HIMapNet \citep{himapnet} decouples hybrid queries into point queries and element queries, and send them into two self-attention mechanism to separately enhance point modeling information and element modeling information. Following, a point-element interactor is applied to query information between points and elements. However, they overlook the relations between points and between elements. GeMap \citep{gemap} enhances the representation of map elements by incorporating a translation- and rotation-invariant representation that effectively leverages the geometry of map elements, and encodes the local structures of map features by displacement vectors. A geometric loss based on angle and magnitude clues is designed, which is robust to rigid transformations of driving scenarios. MGMapNet \citep{mgmapnet} takes element queries, point queries and reference points as input. It dynamically samples point queries directly from Bird’s-Eye View (BEV) features, and integrates them with element queries and reference points, enhancing the geometric accuracy of predicted points; by updating element queries with sampled point features, the overall category and shape information of road elements are effectively captured.

\section{Method}
\subsection{Architecture Overview}

Fig. \ref{fig:framework} presents the overall architecture of SuperMapNet, which mainly consists of a semantic-geometric coupling (SGC) module for the generation of BEV features and a point-element coupling (PEC) module for the classification and localization of map elements.

\textbf{Semantic-Geometric Coupling based BEV Feature Generation.} This module takes both camera images and LiDAR point clouds as inputs, and aims to output fused BEV features with both semantic and geometric information in long ranges. For camera images with semantic information, Swin Transformer \citep{swint} is used as the shared backbone to encode multi-view image features in perspective views; and then, features in perspective views are concatenated and transformed into a unified BEV space by a deformable Transformer \citep{deformabledetr} with geometry priors of cameras. For LiDAR point clouds with accurate geometric information, the generation of BEV features is much simpler than cameras. Original point clouds are first down-sampled to reduce the numbers,following dynamic voxelization in the XOY plane with PointPillars \citep{pointpillar} to generate features of LiDAR point clouds in a BEV space. Secondly, a semantic-geometric coupling (SGC) module is applied, where
a cross-attention based synergy enhancement sub-module is used to capture the complementarity between camera BEV features and LiDAR BEV features, and a flow-based disparity alignment sub-module is used to reduce the coordinate misalignment of different modalities. Finally, a concatenation is used to generate fused BEV features with both rich semantic and geometric information in long ranges.

\textbf{Point-Element Coupling based Map Element Classification and Localization.} Following PivotNet \citep{pivotnet}, our SuperMapNet models map elements in the form of ordered keypoint sets, as the keypoint-based representation can optimize storage efficiency and maintain the construction accuracy. This module takes fused BEV features, point queries and element queries as input, and aims to output locations and classifications of map elements. A point-element coupling (PEC) module is first applied to capture the relations between local and global modeling information, geometric and semantic modeling information. The PEC module includes interactions at three levels, where Point2Point interaction learns local geometric information between points of the same element and of each point, Element2Element interaction learns relation constraints between different elements and semantic information of each elements, and Point2Element interaction learns complement element information for its constituent points. Finally, features with both local and global modeling information, and geometric and semantic modeling information are sent to three decoders of different tasks, a class head to learn elements’ classes, a keypoint head with a dynamic matching module \citep{pivotnet} to regress keypoint coordinates and orders, and a mask head to predict masks of each element.

\begin{figure*}[tp]
    \centering
    \includegraphics[width=\linewidth]{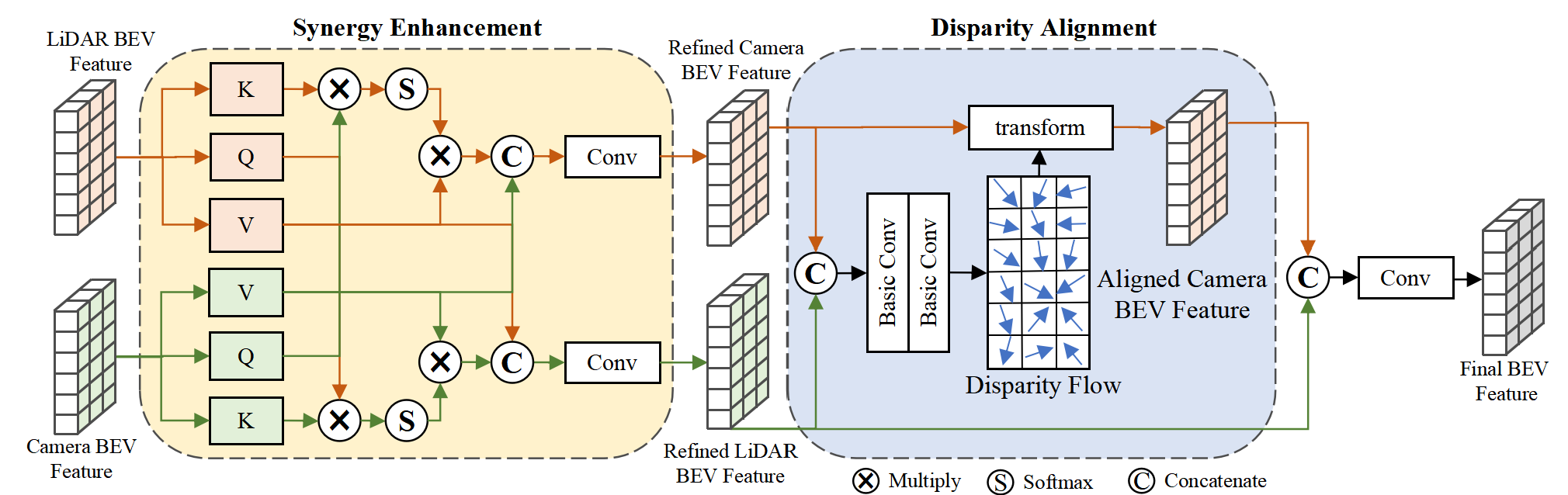}
    \caption{Semantic and geometric coupling (SGC) module. The synergy enhancement is aimed to mining the relations and complementarity between camera BEV featues and LiDAR BEV features, and the disparity alignment is aimed to reduce the coordinate errors between two sensors before concatenation.}
    \label{fig:joint}
\end{figure*}

\subsection{Semantic-Geometric Coupling (SGC) Module}
The perception abilities of LiDAR point clouds and camera images are of different strengths and weaknesses. LiDAR point clouds can provide accurate 3D geometric information, but suffer from disorder and sparsity with an effective range; camera images can capture abundant semantic information about the environment in long ranges, but lack accurate 3D geometric depth information \citep{superfusion}. Multi-modal fusion with both LiDAR point clouds and camera images can effectively complement each other, generating features with rich semantic and geometric information in a long range. However, due to the  inherent complementarity and differences between multi-modal features and the coordinate misalignment between different sensors, directly concatenating features from different modalities results in a low construction accuracy. Therefore, as shown in Fig. \ref{fig:joint}, a semantic-geometric coupling (SGC) module is proposed to fuse features in different modalities, with the consideration of synergies and disparities between different modalities.


\textbf{Synergy Enhancement.} For synergies of camera BEV features and LiDAR BEV features, a cross-attention based enhancement module is proposed to enrich semantic information and fill the feature holes in the distance of LiDAR BEV features, and simultaneously add accurate 3D geometric information to the camera BEV features. For the BEV features $B_{mod}$ of each modality, as listed in follows, three MLP layers are first used to obtain its query $Q_{mod}$, key $K_{mod}$, and value $V_{mod}$, respectively, where $mod$=$\{cam, lidar\}$. 

\begin{equation}
    \centering
     Q_{mod} =MLP(B_{mod})
\end{equation}
\begin{equation}
    \centering
     K_{mod} =MLP(B_{mod})
\end{equation}
\begin{equation}
    \centering
     V_{mod} =MLP(B_{mod})
\end{equation}

Secondly, as listed in Eq. \ref{eq:0} and Eq. \ref{eq:1}, attention matrix $A_{cam2lidar}$ from camera BEV features to LiDAR BEV features is derived from the softmax normalization of the inner product between query $Q_{cam}$ and key $K_{lidar}$ from different modalities, while attention matrix $A_{lidar2cam}$ is obtained from $Q_{lidar}$ and $K_{cam}$. Then, complementary information $C_{cam}$ of camera BEV features is obtained by multiplying attention matrix $A_{cam2lidar}$ with its corresponding value $V_{lidar}$ of another modality, while complementary information $C_{lidar}$ of LiDAR is obtained by multiplying $A_{lidar2cam}$ with $V_{cam}$. Finally, as listed in Eq. \ref{eq:4}, the original value $V_{mod}$ and complementary information $C_{mod}$ of each modality are concatenated and sent to a basic convolution block to learn refined BEV features $B_{mod}$ of each modality with both semantic and geometric information.

\begin{equation}
    \centering
     A_{cam2lidar} =softmax\left(\frac{Q_{cam}K_{lidar}^T}{\sqrt{d_k}}\right) 
    \label{eq:0}
\end{equation}
\begin{equation}
    \centering
     A_{lidar2cam} =softmax\left(\frac{Q_{lidar}K_{cam}^T}{\sqrt{d_k}}\right) 
    \label{eq:1}
\end{equation}
\begin{equation}
    \centering
     C_{cam}= A_{cam2lidar}V_{lidar}
    \label{eq:2}
\end{equation}
\begin{equation}
    \centering
     C_{lidar}= A_{lidar2cam}V_{cam}
    \label{eq:3}
\end{equation}
\begin{equation}
    \centering
    B_{mod}= conv(concat(V_{mod}, C_{mod}))
    \label{eq:4}
\end{equation}
where $mod$=$\{cam, lidar\}$, and ${d_k}$ is a scaling factor.

\textbf{Disparity Alignment.} Due to the disparities between two modalities caused by sensor errors, directly concatenating BEV features of two modalities leads to low accuracy. Thus, a flow-based disparity alignment module is adopted to register the refined camera BEV features to the refined LiDAR BEV features, since the pose accuracy of the LiDAR is always higher compared to the cameras. 

The refined BEV features $B_{cam}\in \mathbb{R}^{C \times H \times W}$ and $B_{lidar}\in \mathbb{R}^{C \times H \times W}$ individually of the camera and the LiDAR obtained in the synergy enhancement module are first concatenated and sent to several basic convolution blocks to obtain a disparity flow $(\Delta_h, \Delta_w) \in \mathbb{R}^{2\times H \times W}$. 

\begin{equation}
    \centering
    (\Delta_h, \Delta_w)= conv(concat(B_{cam}, B_{lidar}))
    \label{eq:5}
\end{equation}

Then, the coordinates of the refined camera BEV features $B_{cam}$ are calibrated by adding the disparity flow $\Delta$ to original coordinates and resampling to generate the aligned camera BEV features $B'_{cam} \in \mathbb{R}^{C \times H \times W}$. The coordinate alignment function of camera BEV features is defined as follows:

\begin{equation}
    \centering
    weight_{h'}=\max\left(0, 1 - \left|w + \Delta_w - \omega \right|\right)
    \label{eq:6}
\end{equation}
\begin{equation}
    \centering
    weight_{w'}=\max\left(0, 1 - \left| h + \Delta_h - h \right|\right)
    \label{eq:7}
\end{equation}
\begin{equation}
    \centering
    B'_{h, w} = \sum_{h'=1}^H \sum_{w'=1}^W B_{h', w'} \cdot weight_{h'} \cdot weight_{w'} \
    \label{eq:8}
\end{equation}
Where a bilinear interpolation kernel is used to sample features on position $(w+ \Delta_w, h+ \Delta_h)$.

After disparities of camera BEV features have been reduced, the LiDAR BEV features $B_{lidar}$ and the aligned camera BEV features $B_{cam}'$ are concatenated and input into a basic convolution to generate the fused BEV features $B \in \mathbb{R}^{C \times H \times W}$ with both semantic and geometric information.

\begin{figure*}[tp]
    \centering
    \includegraphics[width=\linewidth]{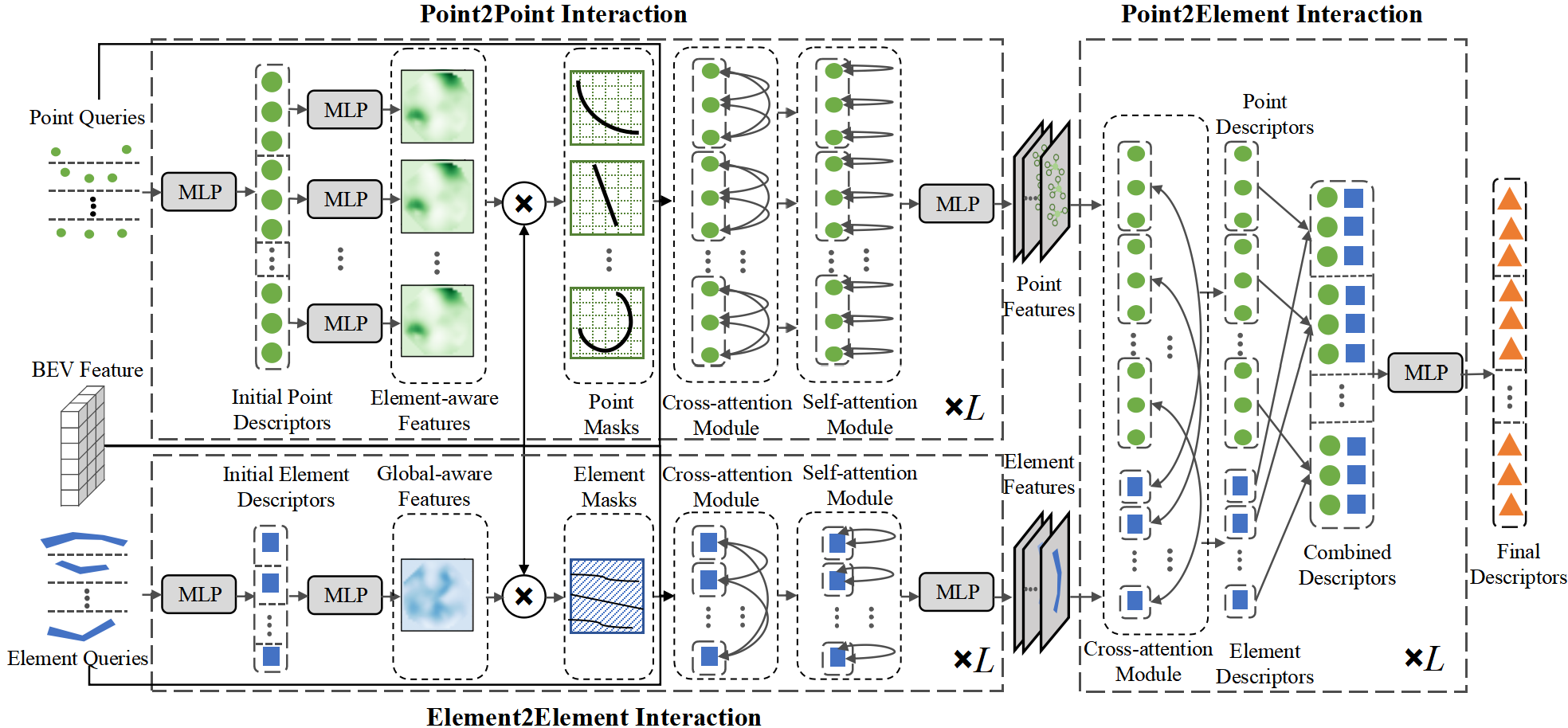}
    \caption{Point and element coupling (PEC) by interactions at three levels, where Point2Point interaction for local information learning between points of the same element and of each point, Element2Element interaction for relation constraints and semantic information learning between different elements and of each elements, and Point2Element interaction for complement element-level information learning of its constituent points.}
    \label{fig:point}
\end{figure*}

\subsection{Point-Element Coupling (PEC) Module} 
For map elements modeled as ordered keypoint sets, there are three levels of information: (1) point-level information, indicating the local coordinates of each point and geometric relationships between points of the same element; (2) element-level information, representing the overall shape and semantic category of each element and relationships between neighbor elements; (3) information between the element and affiliated points, where information of the element provides global constraints and semantic information ton affiliated points, and affiliated points provide specific detail refinements to their element. The three levels of information work in concert with each other. Only utilizing one or two levels of information tends to cause element failures, such as erroneous shapes or entanglement between elements. Thus, as shown in Fig. \ref{fig:point}, we proposed a point-element coupling (PEC) module to fully couple the local and global, semantic and geometric information of points and elements, which is composed of interactions at three levels, Point2Point, Element2Element, Point2Element. 

\textbf{Point2Point Interaction} aims to fully learn both external geometrical relations between points of the same element and internal local information of each point. For each element $m=\left\{1, ..., M\right\}$ in the local map, this module takes the final BEV features $B \in \mathbb{R}^{C \times H \times W}$ and a set of learnable point queries $\left\{Q_{m, n}\right\}_{n=1}^{N} \in \mathbb{R}^{2 \times N}$ as inputs, and outputs their corresponding descriptors $\left\{D_{m, n}\right\}_{n=1}^{N} \in \mathbb{R}^{C \times N}$. 

This module contains $L$ layers, and each layer contains a cross-attention sub-module for external-point information learning, a self-attention sub-module for internal-point information learning, and a feed forward network (FFN) for final point feature generation. In $l$-th layer, a set of initial point descriptors $\left\{{D_{m, n}}^{l-1}\right\}_{n=1}^{N}$ of learnable point queries $\left\{Q_{m, n} \right\}_{n=1}^{N}$ of the element $m$ is concatenated and fed into a multilayer perception to learn element-aware features ${F_{element-aware}}^l \in \mathbb{R}^C$. Then, the element-aware features ${F_{element-aware}}^l \in \mathbb{R}^C$ and the BEV features $B \in \mathbb{R}^{C \times H \times W}$ are multiplied to obtain point masks ${M_{point}}^l \in \mathbb{R}^{H \times W}$. 

\begin{equation}
    \centering
    {F_{element-aware}}^l = MLP(Concat(D_{m,1}, D_{m,2},...,D_{m,N}))
    \label{eq:9}
\end{equation}
\begin{equation}
    \centering
    {M_{point}}^l = \sum_{c=1}^{C} F_{element-aware}(c)*B(c, :,:)
    \label{eq:10}
\end{equation}

A cross-attention mechanism is applied to update external-point information of point descriptors $\left\{{D_{m, n}}^{l} \right\}_{n=1}^{N}$ in the $l$-th layer from the BEV features $B$ and the point descriptors  $\left\{{D_{m, n}}^{l-1} \right\}_{n=1}^{N}$ in the former layer, along with point masks ${M_{point}}^l$ and point position embedding, further enhancing geometric relationships between points of the same element. Following, a self-attention sub-module is applied to learn the internal-point information of point descriptors $\left\{{D_{m, n}}^l \right\}_{n=1}^{N}$. Finally, an FFN sub-module is adopted to generate the final point descriptors $\left\{{D_{m, n}}^l\right\}_{n=1}^{N}$.

\textbf{Element2Element Interaction} aims to fully learn the overall shape and semantic information of each elements and relation constraints between different elements. For all elements of a local map, this module takes the final BEV features $B \in \mathbb{R}^{C \times H \times W}$ and a set of learnable element queries $\left\{Q_{m}\right\}_{m=1}^{M}$ as inputs, and outputs corresponding element descriptors $\left\{D_{m}\right\}_{m=1}^{M}$. 

This module contains $L$ layers, each layer comprising a cross-attention sub-module for external-element information learning, a self-attention sub-module for internal-element information learning, and a feed forward network (FFN) for final element feature generation. In $l$-th layer, a set of initial element descriptors $\left\{{D_{m}}^{l-1}\right\}_{m=1}^{M}$ of learnable element queries $\left\{Q_{m}\right\}_{m=1}^{M}$ is concatenated and fed into a multilayer perception to learn the global-aware features ${F_{global-aware}}^l \in \mathbb{R}^C$. Then, the global-aware features ${F_{global-aware}}^l$ and the BEV features $B \in \mathbb{R}^{C \times H \times W}$ are multiplied to obtain element masks ${M_{element}}^l \in \mathbb{R}^{H \times W}$. 

\begin{equation}
    \centering
    {F_{global-aware}}^l = MLP(Concat(D_{1}, D_{2},...,D_{M}))
    \label{eq:9}
\end{equation}
\begin{equation}
    \centering
    {M_{element}}^l = \sum_{c=1}^{C} F_{global-aware}(c)*B(c, :,:)
    \label{eq:10}
\end{equation}

Masks ${M_{element}}^l$ are subsequently used in the cross-attention layer, along with the BEV features $B$ and element descriptors $\left\{{D_{m}}^{l-1}\right\}_{m=1}^{M}$ in the former layer, to learn overall external-element information, further enhancing semantic relationships between all elements in the same local map. Following, a self-attention layer and an FFN are adopted to learn internal-element information and generate final element descriptors $\left\{{D_{m}}^l\right\}_{m=1}^{M}$, respectively.

\textbf{Point2Element Interaction} aims to use the overall shape and semantic knowledge of an element to complement point-level information of its constituent points, thereby integrating global information while considering details. This module takes the descriptors $D_{m}$ of the element $m$ and the descriptors $\left\{D_{m, n}\right\}_{n=1}^{N}$ of its constituent points $\left\{Q_{m, n}\right\}_{n=1}^{N}$ as input, and aims to update both point descriptors and element descriptors.

This module contains $L$ layers. In the $l$-th layer, a cross-attention sub-module with position embedding is applied for information communication between descriptors ${D_{m}}^l$ of the element $m$ and descriptors $\left\{{D_{m, n}}^l\right\}_{n=1}^{N}$ of its constituent points. The update step of two descriptors in $(l+1)$-th layer can be expressed as follows:

\begin{equation}
    \centering
     {Q_{m, i}}^{l} = MLP({D_{m, i}}^{l}),{K_{m, i}}^{l} = MLP({D_{m, i}}^{l})
\end{equation}
\begin{equation}
    \centering
     {Q_{m}}^{l} = MLP({D_{m}}^{l}),{K_{m}}^{l} = MLP({D_{m}}^{l})
\end{equation}
\begin{equation}
    \centering
    {D_{m, i}}^{l+1} = {D_{m, i}}^{l} + softmax\left(\frac{{Q_{m, i}}^{l}({K_{m}}^{l})^T}{\sqrt{d_k}}\right){D_{m}}^{l}
\end{equation}
\begin{equation}
    \centering
    {D_{m}}^{l+1} = {D_{m}}^{l} + softmax\left(\frac{{Q_{m}}^{l}({K_{m, i}}^{l})^T}{\sqrt{d_k}}\right){D_{m, i}}^{l}
\end{equation}
where ${D_{m, i}}^l$ is the $i$-th point's descriptor of the element $m$ in $l$-th layer, ${D_{m}}^l$ is the descriptor of the element $m$ in $l$-th layer. $Q$ and $K$ are obtained by two fully-connected layers, respectively. ${d_k}$ is a scaling factor.

\section{Experiments}
\label{sec:experiments}

\subsection{Experimental Settings }
\begin{figure}[]
    \centering
    \subfloat[Boston Seaport.]{
        \includegraphics[width=0.24\textwidth, height = 80pt]{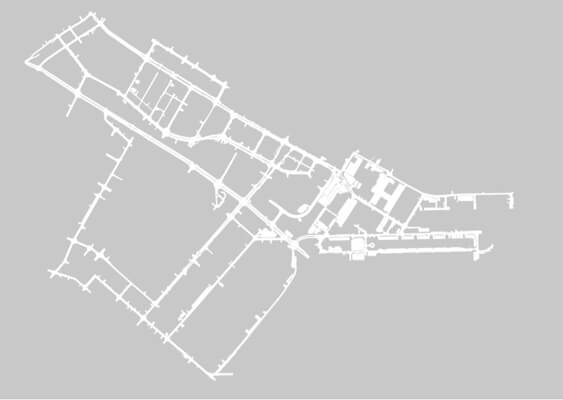}
   }
     \subfloat[Singapore Holland Village.]{
         \includegraphics[width=0.24\textwidth, height = 80pt]{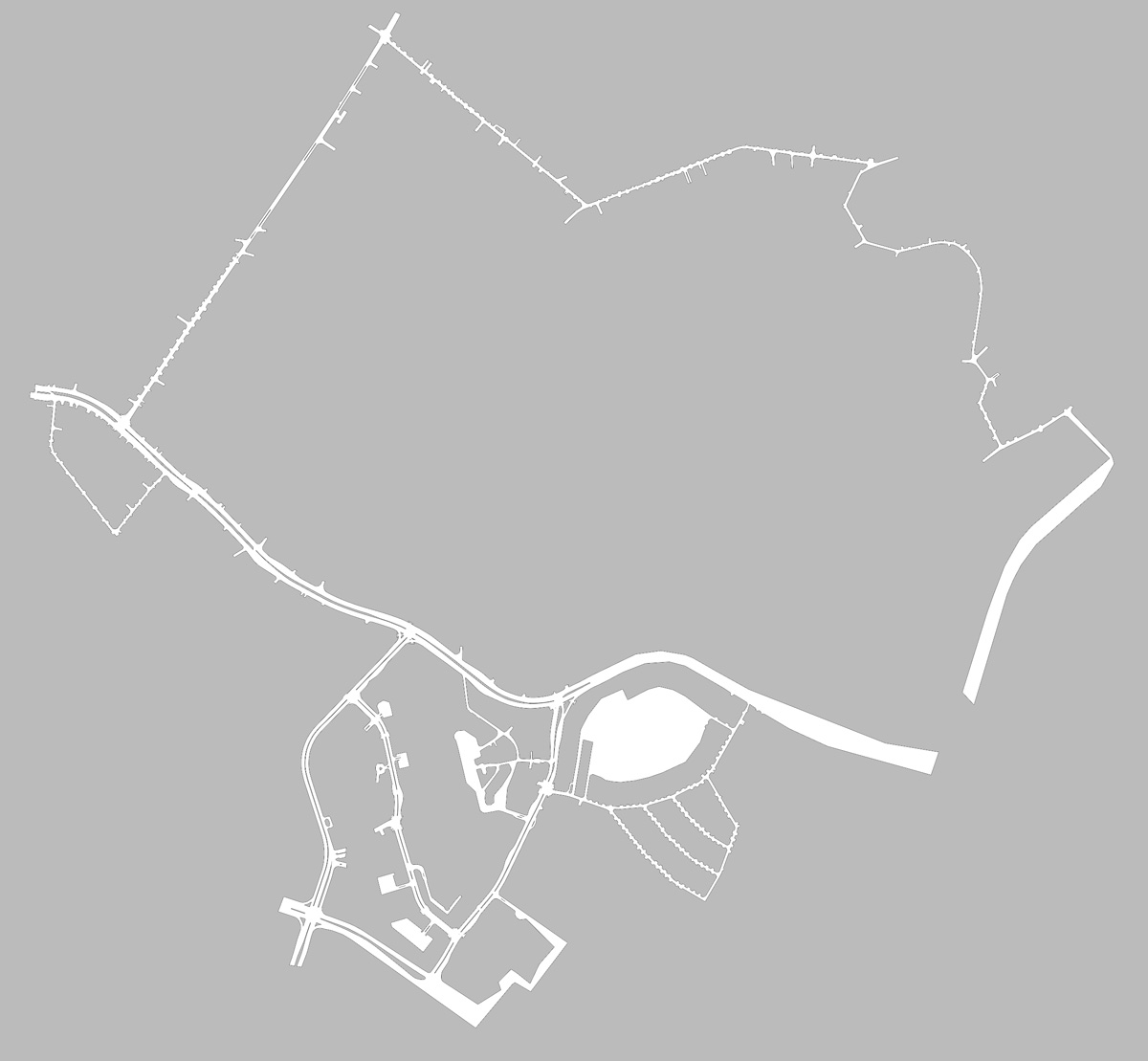}
   }
    \hfill
    \subfloat[Singapore Queenstown.]{
        \includegraphics[width=0.24\textwidth, height = 80pt]{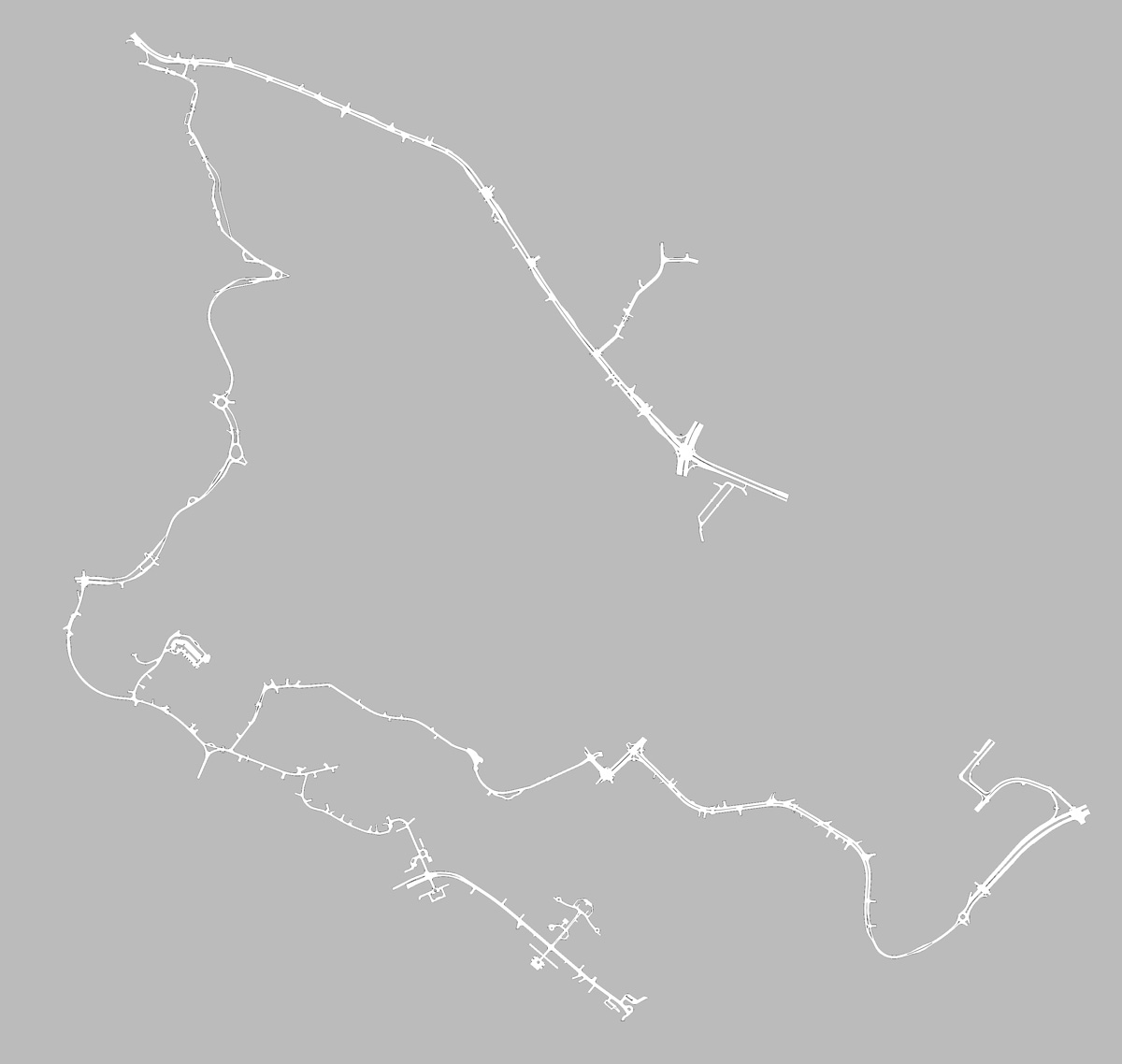}
   }
   \subfloat[Singapore One North.]{
         \includegraphics[width=0.24\textwidth , height = 80pt]{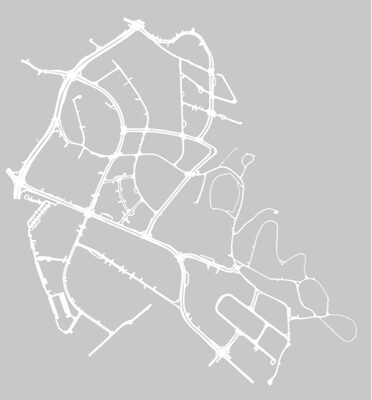}
   }
    \caption{Global view of nuScenes dataset. }
    \label{fig:nuscenes}
\end{figure}

\textbf{NuScenes Dataset.} As shown in Fig. \ref{fig:nuscenes}, nuScenes dataset \citep{nuscenes} comprises 1000 scenes collected from Boston Seaport and Singapore’s One North, Queenstown and Holland Village districts, which are renowned for their dense traffic and highly challenging driving conditions. Each scene lasts around 20 seconds and is annotated at a frequency of 2Hz. Each sample includes 6 RGB images from surrounding cameras and point clouds from LiDAR sweeps. Following previous methods \citep{pivotnet, bemapnet}, 700 scenes with 28130 samples are used for training and 150 scenes with 6019 samples are used for validation and 6019 samples are used for testing. For a fair comparison, we focus on three categories of map elements, including road boundaries, lane dividers, and pedestrian crossings. 

\textbf{Argoverse2 Dataset.} It \citep{argoverse2} contains 1000 logs from six different cities in the United States, Austin, Detroit, Miami, Palo Alto, Pittsburgh, and Washington D.C., recording from different seasons, weather conditions, and various times of the days. Each log contains 15 seconds of 20Hz RGB images from 7 ring cameras and 2 stereo cameras, 10Hz LiDAR sweeps, and a 3D vectorized map. Following the previous work \citep{himapnet}, 700 logs with 108,972 samples are used for training, 150 logs with 23,542 samples for validation, and another 150 logs with 23,542 samples for testing. We focus on the same map element categories as nuScenes dataset. 

\textbf{Evaluation Metric.} Following previous methods \citep{pivotnet, bemapnet, himapnet, vectormapnet}, the common average precision (AP) based on the Chamfer distance is adopted as the evaluation metric, where a prediction is treated as true positive (TP) only if the distance between the prediction and the ground-truth is less than a threshold. Since existing methods used different AP thresholds for evaluation, we set two different threshold sets \{0.2, 0.5, 1.0\}$m$ and \{0.5, 1.0, 1.5\}$m$ corresponding to hard and easy settings. For each setting, the final AP result is calculated by averaging across three thresholds and all classes.

\textbf{Implementation details.} With the ego-car as the center, the perception ranges of the local map are set to [\textminus15.0 $m$, 15.0 $m$] in the X-axis and [\textminus60.0 $m$, 60.0 $m$] in the Y-axis. We set the image size of NuScenes dataset to 512 × 896 and Argoverse2 dataset to 384 × 512, and voxelize the LiDAR point clouds of both datasets in 0.15 $m$. We set the size of BEV features to 100 × 25. The maximum numbers $M$ of element type, such as lane dividers, pedestrian crossings, and road boundaries are set to $\{20, 25, 15\}$, and the maximum numbers $N$ of modeled keypoints of each element type are set to $N = \{10, 2, 30\}$, respectively. We perform all the experiments on a machine equipped with 4 Tesla V100-DGXS-32G GPUs. During the training phase, all GPUs are utilized, whereas only a single GPU is employed for inference. NuScenes dataset is trained for 30 epochs, and Argoverse2 dataset for 6 epochs, using the AdamW optimizer and an exponential scheduler, with a learning rate of 0.0001 and a weight decay of 0.0001. The batch size of training is set to 4.

\begin{table*}[t]
\caption{Comparison to the SOTAs on nuScenes dataset. Among all methods, the best results are in bold and the second in underline, and gains in red are calculated based on the best and the second results. The results of the comparative methods are referenced in their paper. “-” means that the corresponding results are not available. FPS is measured on NVIDIA RTX 3090 GPU with batch size of 1. “C” denotes the use of camera, and “L” denotes the use of LiDAR.}
\resizebox{\linewidth}{!}
{
\begin{tabular}{ccc|cccc|cccc|cc}
\toprule
$Method$      & $Modality$         & $Epoch$            & $AP_{ped}$            & $AP_{div}$           & $AP_{bou}$          & $mAP$       & $AP_{ped}$    
& $AP_{div}$             & $AP_{bou}$                 & $mAP$   & $FPS$ & $Params.$ \\
             &          &           & \multicolumn{4}{c|}{$Hard$:\{0.2, 0.5, 1.0\}m}        & \multicolumn{4}{c}{$Easy$:\{0.5, 1.0, 1.5\}m}            \\
\midrule
BeMapNet \citep{bemapnet}                        & C                                              & 30  & 42.2 & 49.1 & 39.9 & 43.7 & 61.2 & 64.4 & 61.7 & 62.4                       &4.2    &73.8            \\
MapQR \citep{mapqr}  &C    &110 &- &-   &-   &-   &74.4 &70.1 &73.2 &72.6    & 11.9            & 125.3\\
MapTR \citep{maptr}	&C		&110	&31.4	&40.5	&35.5	&35.8	&55.8	&60.9	&61.1	&59.3   &15.1 &35.9\\
MapTRv2 \citep{maptrv2}	&C		&110	&43.6	&49.0	&43.7	&45.4	&68.1	&68.3	&69.7	&68.7  &14.1 &40.3 \\
PivotNet \citep{pivotnet}	&C		&110	&43.4	&53.6	&\underline{50.5}	&49.2	&62.6	&68.0	&69.7	&66.8  &9.2  &44.8\\
MGMapNet \citep{mgmapnet}     &C  &110   &-  &-  &-   &-  &64.4   &67.6  & 67.7   &66.5   &11.6   &70.1  \\
 HIMapNet \citep{himapnet}                       & C                                      & 110          & \underline{47.3} &\underline{57.8} &49.6 &\underline{51.6} &71.3 &\underline{75.0} &74.7 &73.7         &11.4  &63.2                 \\
\midrule

MapTR \citep{maptr}                           & C+L                                         & 24                          &  -                            &   -                           & -                             & -                             & 55.9 & 62.3 & 69.3 & 62.5   &6.0  &- \\
MapTRv2 \citep{maptrv2}                        & C+L                                        & 24                          &   -                           &  -                            & -                             & -                             & 65.6                         & 66.5                         &74.8                         & 69.0                    &5.8 &-                                  \\
MGMapNet \citep{mgmapnet}     &C+L  &24   &-  &-  &-   &-  &67.7   &71.1 & 76.2   &71.7   &4.8   &-  \\
HDMapNet \citep{hdmapnet}                       & C+L                                         & 30                          & 7.1                          & 28.3                         & 32.6                         & 22.7                         & 16.3 & 29.6 & 46.7 &31.0            &0.5          &69.8                  \\
InsMapper \citep{insMapper}     &C+L       &30  &- &-  &-   &-  &56.0   &63.4   &71.6  &63.7  &-  &-\\
VectorMapNet \citep{vectormapnet}                   & C+L                                            & 110                         & -                            & -                            & -                            & -                            & 37.6 & 50.5 &47.5 & 45.2                       &-  &-                    \\
GeMap \citep{gemap}  &C+L      & 110   &-  &-   &-    &-   & 69.8    &68.0    &73.4     & 70.4  &6.8 &- \\
ADMap \citep{admap} & C+L  &110 &-  &-   &-    &-  &66.5 &71.2 &76.9  &71.5  &5.8  &-\\
HIMapNet \citep{himapnet}                       & C+L                                      & 110          & - &- &- &- &\underline{77.0} &74.4 &\textbf{82.1} &\underline{77.8}         &-  &-                 \\
\midrule
SuperMapNet                            & C+L                                        & 30                          & \textbf{70.3}\textcolor{red}{(+23.0)}              & \textbf{73.8}\textcolor{red}{(+16.0)}                & \textbf{55.5}\textcolor{red}{(+5.0)}                         & \textbf{66.5}\textcolor{red}{(+14.9)}                & \textbf{88.8}\textcolor{red}{(+11.8)}                & \textbf{90.9}\textcolor{red}{(+15.9)}                & \underline{80.2}                          & \textbf{86.6}\textcolor{red}{(+8.8)}  &5.0  &69.7   \\
\bottomrule                                              
\end{tabular}
}
\label{tab:1}
\end{table*}

\begin{table*}[]
\caption{Comparison to the SOTAs on Argoverse2 dataset. Among all methods, the best results are in bold and the second in underline, and gains in red are calculated based on the best and the second results. The results of the comparative methods are referenced in their paper. “-” means that the corresponding results are not available. “C” denotes the use of camera, and “L” denotes the use of LiDAR.}
\resizebox{\linewidth}{!}
{
\begin{tabular}{ccc|cccc|cccc}
\toprule
$Method$      & $Modality$            & $Epoch$            & $AP_{ped}$            & $AP_{div}$           & $AP_{bou}$          & $mAP$       & $AP_{ped}$    
& $AP_{div}$             & $AP_{bou}$                 & $mAP$  \\
             &          &                & \multicolumn{4}{c|}{$Hard$:\{0.2, 0.5, 1.0\}m}        & \multicolumn{4}{c}{$Easy$:\{0.5, 1.0, 1.5\}m}            \\
\midrule
MapTR \citep{maptr}        & C             & 6     & 28.3       & 42.2       & 33.7       & 34.8       & 54.7       & 58.1       & 56.7       & 56.5       \\
MapTRv2 \citep{maptrv2}     & C              & 6     & 34.8       & 52.5       & 40.6       & 42.6       & 63.6       & 71.5      & 67.4       & 67.5         \\
MapVR \citep{mapvr}       & C             & 6     & -          & -          & -          & -          & 54.6       & 60.0       & 58.0       & 57.5          \\ 
PivotNet \citep{pivotnet}    & C           & 6     & 30.6       & 48.0       & \underline{44.5}       & 41.0       & -          & -          & -          & -           \\
HIMapNet \citep{himapnet}     & C             & 6     & \underline{39.9}       & 53.4       & 44.3       & \underline{45.8}       & 69.0       & 69.5       & 70.3       & 69.6         \\
MapQR \citep{mapqr}        & C             & 6     & 36.5       & \underline{56.3}       & 42.5       & 45.1      & 64.3       & 72.3       & 68.1       & 68.2          \\
InsMapper \citep{insMapper}  &   C   &6  &- &-  &-   & -   & 55.6   &66.6    &62.6     &61.6  \\ 
MGMapNet \citep{mgmapnet}    &C   &6   &-  &-  &-  &- &52.8   &67.5    &68.1   &62.8   \\
VectorMapNet \citep{vectormapnet} & C              & 24    & 18.3       & 33.3       & 20.4       & 24.0       & 38.3       & 36.1       & 39.2       & 37.9      \\ 
GeMap \citep{gemap}   & C     &24   &- &-  &-   & - & 75.7  & 69.2 &70.5  &71.8 \\ 
\midrule
HDMapNet \citep{hdmapnet}     & C+L           & 6     & 9.8        & 19.5       & 35.9       & 21.8       & 13.1       & 5.7        & 37.6       & 18.8       \\
ADMap \citep{argoverse2}  & C+L  &6   &- &-  &-   & -    &75.5 &69.5 &80.5 & 75.2   \\

HIMapNet \citep{himapnet}     & C+L              & 6     & -       & -       & -       & -       & \underline{78.7}       & \underline{75.7}       & \underline{83.3}       & \underline{79.3}        \\
\midrule
SuperMapNet         & C+L       & 6     & \textbf{61.5}\textcolor{red}{(+21.6)}  
& \textbf{69.7}\textcolor{red}{(+13.4)}  & \textbf{61.8}\textcolor{red}{(+17.3)} & \textbf{64.3}\textcolor{red}{(+18.5)} & \textbf{81.8}\textcolor{red}{(+3.1)} & \textbf{79.9}\textcolor{red}{(+4.2)} & \textbf{85.6}\textcolor{red}{(+2.3)}  & \textbf{82.4}\textcolor{red}{(+3.1)}  \\
\bottomrule   
\end{tabular}
}
\label{tab:2}
\end{table*}

\begin{table*}[]
\caption{Accuracy of different element types on nuScenes val set. Gains in red are calculated based on the baseline. }
\resizebox{\linewidth}{!}
{
\begin{tabular}{cc|cccc|cccc|cc}
\toprule
     $SGC$    & $PEC$  & $AP_{ped}$            & $AP_{div}$           & $AP_{bou}$          & $mAP$       & $AP_{ped}$    & $AP_{div}$             & $AP_{bou}$                 & $mAP$    &$FPS$     &$Params.$\\
                              &               & \multicolumn{4}{c|}{$Hard$:\{0.2, 0.5, 1.0\}m}                                 & \multicolumn{4}{c}{$Easy$:\{0.5, 1.0, 1.5\}m}     &   &                                                      \\
\midrule

 &                                                       & 62.2                         & 63.3                         & 46.9                        & 57.5                       & 83.6                         & 83.3                         & 72.1                        & 79.7  &6.0    &56.1\\
        $\checkmark$  &                                                      & 68.6 \textcolor{red}{(+6.4)}                & 66.7 \textcolor{red}{(+3.4)}  & 52 \textcolor{red}{(+5.1)}     & 62.4 \textcolor{red}{(+4.9)}         & 86.2 \textcolor{red}{(+2.6)}  & 85.7 \textcolor{red}{(+2.4)}  & 76.7 \textcolor{red}{(+4.6)}   & 82.9 \textcolor{red}{(+3.2)}                      &5.8    &59.3                           \\
         &$\checkmark$          & 69.1 \textcolor{red}{(+6.9)} & 72.2 \textcolor{red}{(+8.9)}          & 54.3 \textcolor{red}{(+7.4)}    & 65.2 \textcolor{red}{(+7.7)}               & 89.1 \textcolor{red}{(+5.5)}           & 90.5 \textcolor{red}{(+7.2)}        & 79.5  \textcolor{red}{(+7.4)}  & 86.4  \textcolor{red}{(+6.7)}        &5.2  &66.5                                                \\

        $\checkmark$  &$\checkmark$                                                 & 70.3 \textcolor{red}{(+8.1)}     & 73.8 \textcolor{red}{(+10.5)}  & 55.5 \textcolor{red}{(+8.6)}  & 66.5 \textcolor{red}{(+9.0)}    & 88.8 \textcolor{red}{(+5.2)}  & 90.9 \textcolor{red}{(+7.6)}  & 80.2 \textcolor{red}{(+8.1)}   &86.6 \textcolor{red}{(+6.9)}  &5.0   &69.7\\
\bottomrule 
\end{tabular}
}
\label{tab:modules}
\end{table*}

\subsection{Comparisons with SOTAs } 
\textbf{Results on NuScenes}. Tab. \ref{tab:1} shows the performances of our SuperMapNet compared with existing SOTAs on nuScenes dataset under hard and easy settings. SuperMapNet achieves new SOTAs (66.5/86.6 mAP) in both hard and easy settings, with increases of 14.9/8.8 mAP compared to the second-best methods, respectively, especially on lane dividers and pedestrian crossings, with improvements over 10 mAP. Although the accuracy of road boundaries under the easy setting is 1.9 mAP lower than HIMapNet \citep{himapnet}
with multi-modalities, the training epochs (30 epochs) of our SuperMapNet are much shorter than other methods (110 epochs) and the perception range of the generated local maps (120 $m$ in the Y-axis) is much larger than other methods (60 $m$ in the Y-axis), showing the significant superiority of our SuperMapNet in improving the construction accuracy and perception range. As for the inference latency, with 5.0 FPS (Frames Per Second) on an RTX 3090 GPU, SuperMapNet maintains real-time capability while delivering top-tier accuracy. Although the FPS of our SuperMapNet with both camera images and LiDAR point clouds as input is much lower compared to single modality-based construction methods, it is on a superior accuracy-speed balance compared to most multi-modal methods. Moreover, SuperMapNet contains 69.7M parameters, which is 45\% smaller than MapQR \citep{mapqr} with single modality (125.3 M) and is comparable to HIMapNet \citep{himapnet} (63.2 M) and BeMapNet \citep{bemapnet} (73.8 M) both in single modality, indicating efficient design despite its multi-modal input.

\begin{figure}[]
    \centering
    \subfloat[the complex distribution and shape of pedestrian crossings.]{
        \includegraphics[width=0.24\textwidth ]{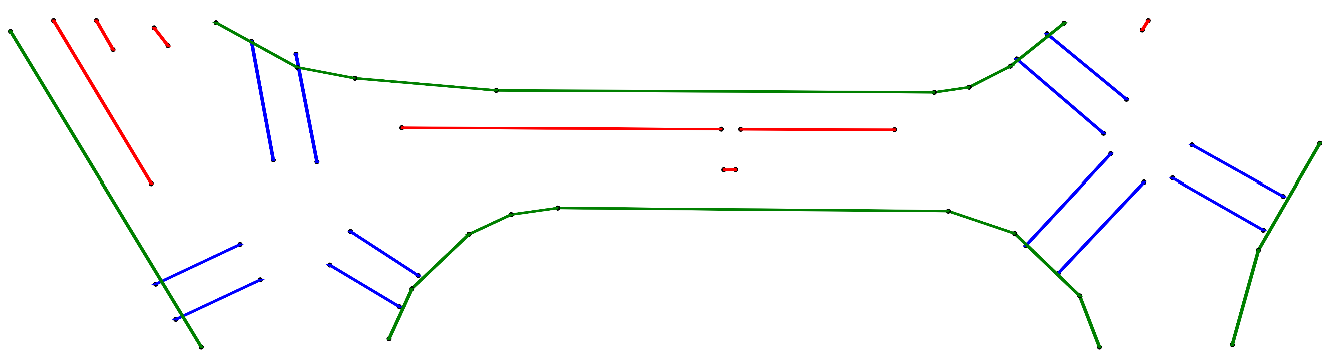}
         \includegraphics[width=0.24\textwidth]{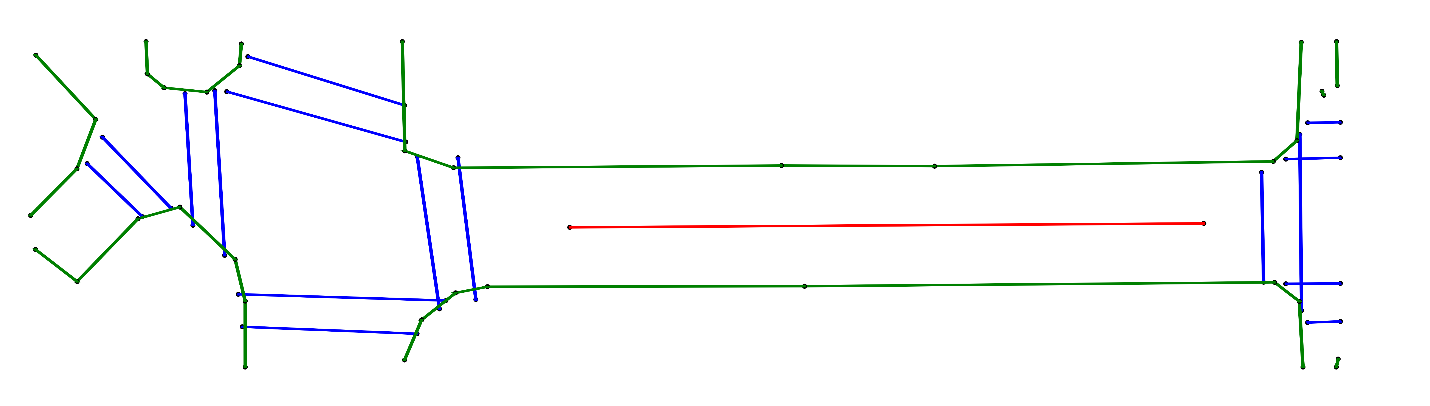}
   }
    \hfill
    \subfloat[repeated labeling between boundaries and dividers, and between dividers.]{
        \includegraphics[width=0.24\textwidth, ]{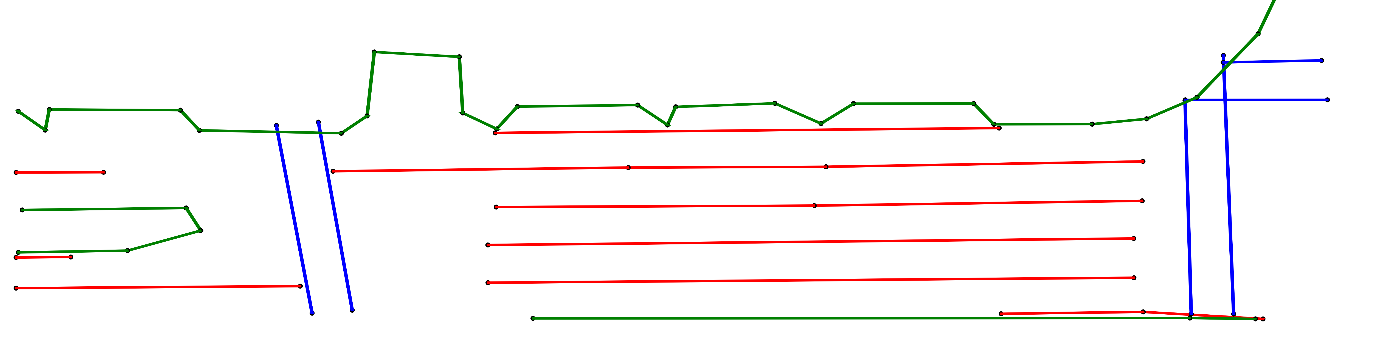}
         \includegraphics[width=0.24\textwidth ]{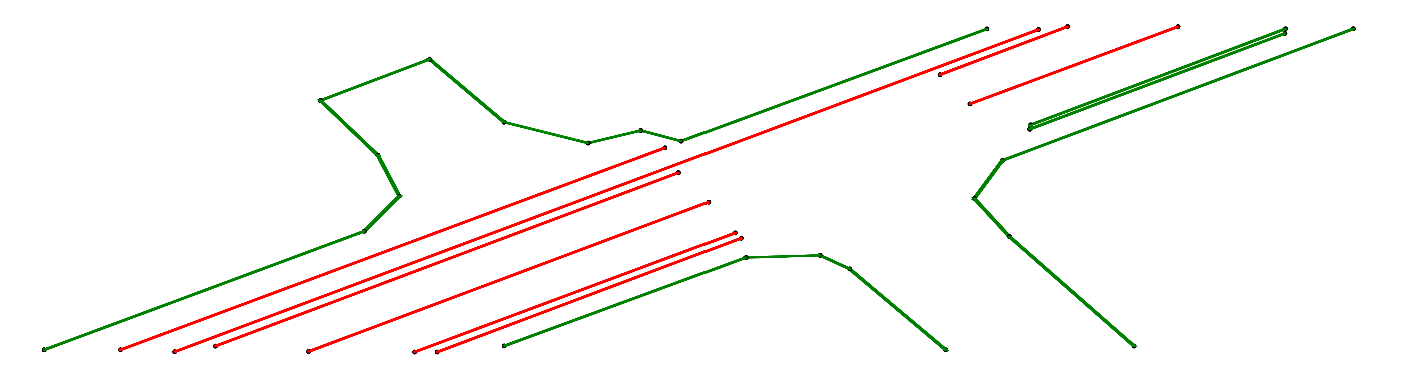}
   }
    \caption{Challenges on Argoverse2 dataset. }
    \label{fig:challengs}
\end{figure}

\textbf{Results on Argoverse2}. Tab. \ref{tab:2} shows the performances of our SuperMapNet with existing SOTAs on Argoverse2 \citep{argoverse2} dataset under hard and easy settings. Due to the fact that the data volume of the Argoverse2 dataset is about four times than that of the NuScenes dataset, most methods only train 6 epochs on the Argoverse2 dataset. In addition, existing methods rarely conduct experiments on the Argoverse2 dataset with multi-modalities as input, due to (1) the larger dataset size imposes higher memory requirements; and (2) the sparse density of LiDAR point clouds on Argoverse2 dataset reduce the advantages of multi-modal fusion. To the best of our knowledge, only three existing methods (HDMapNet \citep{hdmapnet}, ADMap \citep{admap}, and HIMapNet \citep{himapnet}) have performed experiments on Argoverse2 dataset with multi-modalities. It is obvious that our SuperMapNet sets a new SOTA on Argoverse2 dataset by delivering the highest accuracy (18.5/3.1 mAP) in both hard and easy settings among all methods, even though the perception range of SuperMapNet is the twice than other methods. However, compared to nuScenes dataset, the performance of our SuperMapNet shows a slight decrease in accuracy, about 2.2/4.2 mAP in the hard/easy setting, especially in pedestrian crossings (\textminus8.8/\textminus7.0 mAP) and lane dividers (\textminus4.1/\textminus11.0 mAP). This is due to the facts that, as shown in Fig. \ref{fig:challengs}, the distribution and shape of pedestrian crossings is more complex with repeated labeling between boundaries and dividers, and between dividers, which bring great challenges to their construction. Compared to camera-only methods, SuperMapNet achieves 40\% higher than the second-best model under the hard setting, demonstrating the indispensable role of LiDAR in decimeter or centimeter HD map construction. Compared to existing multi-modal methods, SuperMapNet still achieved the highest accuracy on each element type, showing the great potential of our SuperMapNet.

\section{Ablation Study}
\subsection{Accuracy of different element types}

The significant differences in the shapes of different types of elements bring challenges for modeling map elements in a unified representation. Tab. \ref{tab:modules} lists how the proposed SGC module and the PEC module affect the accuracy of different element types. Notably, even without incorporating SGC or PEC modules, our baseline model with multi-modalities already achieves higher accuracy compared to existing SOTAs, where HIMapNet \citep{himapnet} attains an accuracy of 77.8 mAP under the easy setting, while our baseline achieves 79.7 mAP in the same scenario. 

\textbf{SGC module}: Compared to the strategy of directly concatenating in the baseline, adding the SGC module before concatenation achieves 4.9/3.2 mAP higher under the hard/easy setting, with slightly reducing FPS by 0.2) and increasing parameters by 3.2M. This is owing to (1) the cross-attention based synergy enhancement for information exchanging between semantic information and geometric information, which can fill the feature holes in the distance of LiDAR BEV features and enhance the perception ability and range; (2) the flow-based disparity alignment to reduce the coordinate errors between different sensors before concatenation. 

\textbf{PEC module}: Using the PEC module only yields a much higher accuracy (+7.7/+6.7 mAP) compared to using the SGC module only under both hard and easy setting, especially on road boundaries and lane dividers, with an increase of 8.9/7.2 mAP and 7.4/7.4 mAP, respectively. This is because pedestrian crossings are typically modeled by only two keypoints, with limited point information, leading to the Point2Point interation in PEC module working invalidly. In contrast, road boundaries and lane dividers are modeled by more keypoints with rich point information, thus, Point2Point interation for road boundaries and lane dividers is more effective. However, FPS of using the PEC module only drops significantly by 0.8, and parameters grow by +10.4M, as three-levels of interations, Point2Point, Element2elEment and Point2Element, is added.

\textbf{Combination of SGC and PEC module}: Combining both SGC and PEC modules do not fully amplify their advantages, especially under the easy settings where the effect of adding the SGC module or not is not much different (the changes of mAP from 86.4 to 86.6). However, it is still necessary to combine both two modules in the hard settings, which can increase mAP by 1.2, 1.6, and 1.2 on pedestrian crossings, road boundaries, and lane dividers, respectively, compared to using PEC module only.



\subsection{Accuracy of different thresholds}

Varying AP thresholds represent different tolerances for performance evaluation. Considering the application of autonomous driving systems in perception, prediction and planning, HD maps require centimeter-level information for safety, reliability, and intelligence. Thus, improvements under stricter thresholds are more practical and meaningful. Tab. \ref{tab:3} shows how the proposed SGC module and the PEC module affect the accuracy under different thresholds $(0.2m, 0.5m, 1.0m, 1.5m)$. Noting that the baseline already perform well for large thresholds, 82.9 mAP in the threshold of 1.0 $m$ and 88.1 mAP in the threshold of 1.5 $m$.

\textbf{SGC module}: Tab. \ref{tab:3} shows that, the advantage of the proposed SGC module decreases sharply with the increase of threshold. When the threshold is 0.2 $m$, using SGC module can improve 7.0 mAP. However, when the threshold is amplified to 1.0 $m$ and 1.5 $m$, only 2.7 mAP and 1.6 mAP is improved compared to the baseline, respectively. This is because the disparity alignment in the SGC module aims to reduce the coordinate error between different sensors. However, such errors (typically at decimeter or centimeter scales) exert minimal influence on the accuracy of meter-level map reconstruction. Certainly, for applications demanding higher accuracy (e.g., decimeter/centimeter-level map reconstruction), such sensor coordinate alignment becomes critical.

\textbf{PEC module}: Tab. \ref{tab:3} shows that, the PEC module demonstrates consistently greater improvements across all thresholds compared to the SGC module, and achieves more stable performance under the thresholds of 0.2 $m$, 0.5 $m$, and 1.0 $m$, with the improvement exceeding +6.0 mAP over the baseline. This is thanks to that, the interactions at three levels between points and elements to learn local geometric information between points of the same element and of each point, relation constraints between different elements and semantic information of each elements, and complementary information of element for its constituent points, effectively reducing erroneous shapes and entanglement between elements. However, at the 1.5 $m$ threshold, the improvement of the PEC module decreases to +3.7 mAP. This improvement is meaningless, because the tolerance of this threshold (1.5 $m$) is too high, even if erroneous shapes and entanglement between elements occur, it will still be considered as correct construction at the threshold of 1.5 $m$.

\textbf{Combination of SGC and PEC modules}: Tab. \ref{tab:3} shows that, the superiority of combining both SGC and PEC modules is most pronounced under strict thresholds. For instance, it achieves a +3.2 mAP improvement compared to using either SGC or PEC module individually at the threshold of 0.2 $m$. However, this advantage diminishes significantly at larger thresholds, especially at the thresholds of 1.0 $m$ and 1.5 $m$, this advantage basically does not exist, and achieves the same accuracy as only using PEC module, as the SGC module. As mentioned above, considering safety, reliability, and intelligence of autonomous driving systems, evaluations under stricter thresholds are more practical and meaningful. Thus, for applications demanding higher accuracy (e.g., decimeter/centimeter-level map reconstruction), combining both SGC module with PEC module is critical. 

\begin{table}[]
\centering
\caption{Accuracy of different thresholds on nuScenes val set. Gains in red are calculated based on the baseline.}
\resizebox{\linewidth}{!}
{
\begin{tabular}{cc|cccc}
\toprule
     $SGC$    & $PEC$        & $AP_{0.2 m}$       & $AP_{0.5 m}$       & $AP_{1.0 m}$       & $AP_{1.5 m}$       \\
\midrule
                &  &21.4  &68  &82.9   &88.1\\
$\checkmark$                 &   &28.4 \textcolor{red}{(+7.0)}   &73.2 \textcolor{red}{(+5.2)}   &85.6 \textcolor{red}{(+2.7)}    &89.7 \textcolor{red}{(+1.6)}  \\
                &$\checkmark$    &29.3  \textcolor{red}{(+7.9)}   &77.1 \textcolor{red}{(+8.9)}    &89.2 \textcolor{red}{(+6.3)}      &92.8 \textcolor{red}{(+3.7)} 
\\
 $\checkmark$                 &$\checkmark$           &32.5 \textcolor{red}{(+11.1)} & 77.9 \textcolor{red}{(+9.9)}  & 89.2 \textcolor{red}{(+6.3)} & 92.8 \textcolor{red}{(+3.7)} \\
\bottomrule  
\end{tabular}
}
\label{tab:3}
\end{table}

\subsection{Visualization}
\begin{figure*}[tp]
    \centering
    \subfloat{
       \includegraphics[width=0.24\textwidth, height = 160pt]{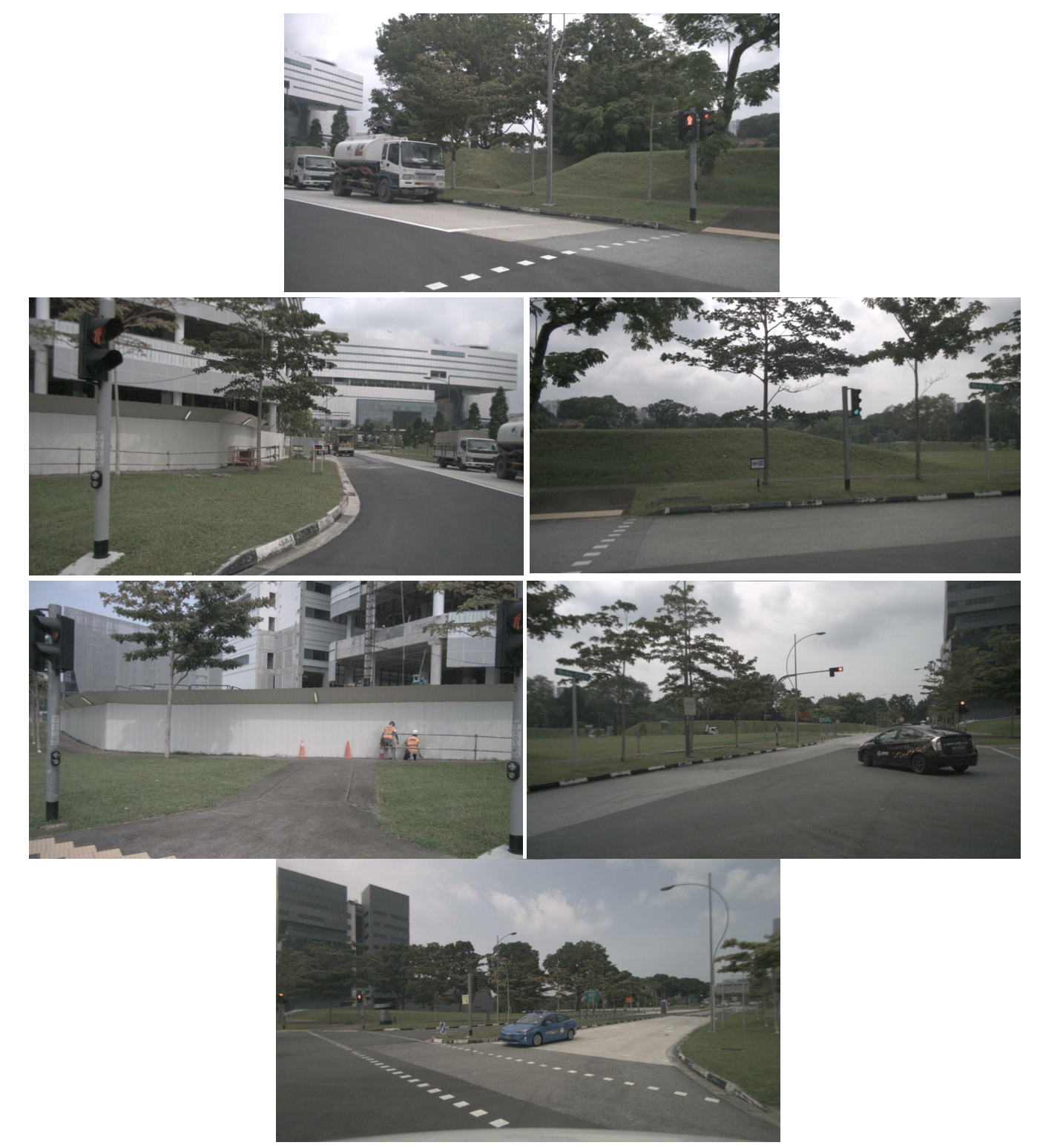}
       \includegraphics[width=0.12\textwidth, height = 160pt]{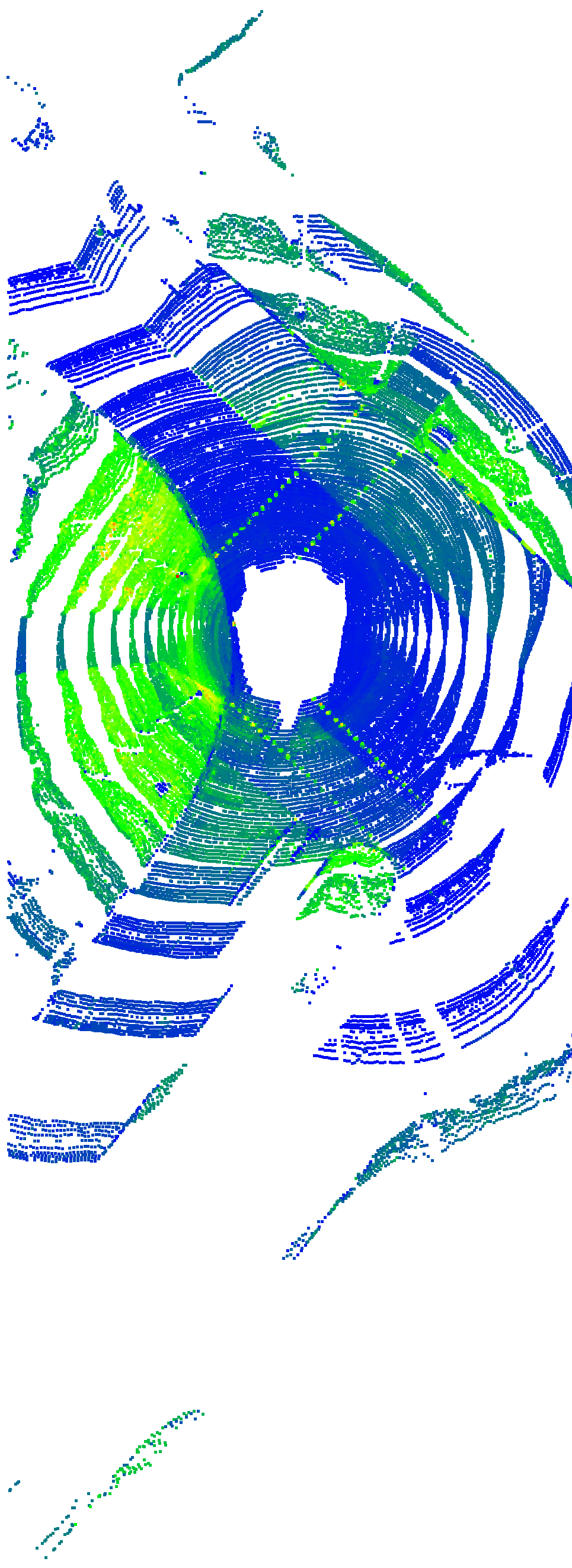}
       \includegraphics[width=0.12\textwidth, height = 160pt]{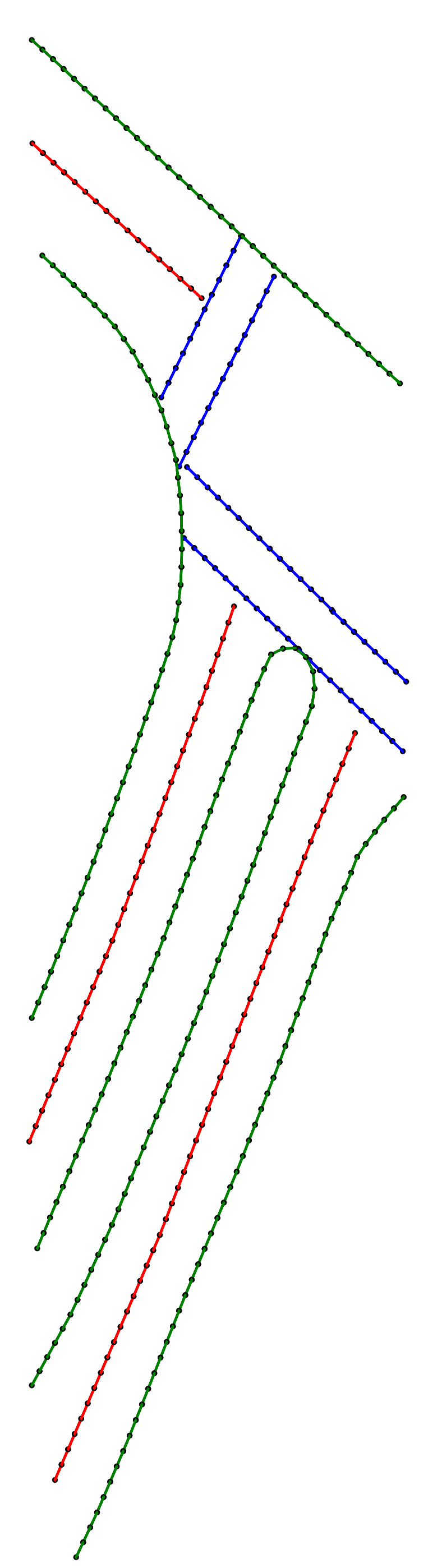}
       \includegraphics[width=0.12\textwidth, height = 160pt]{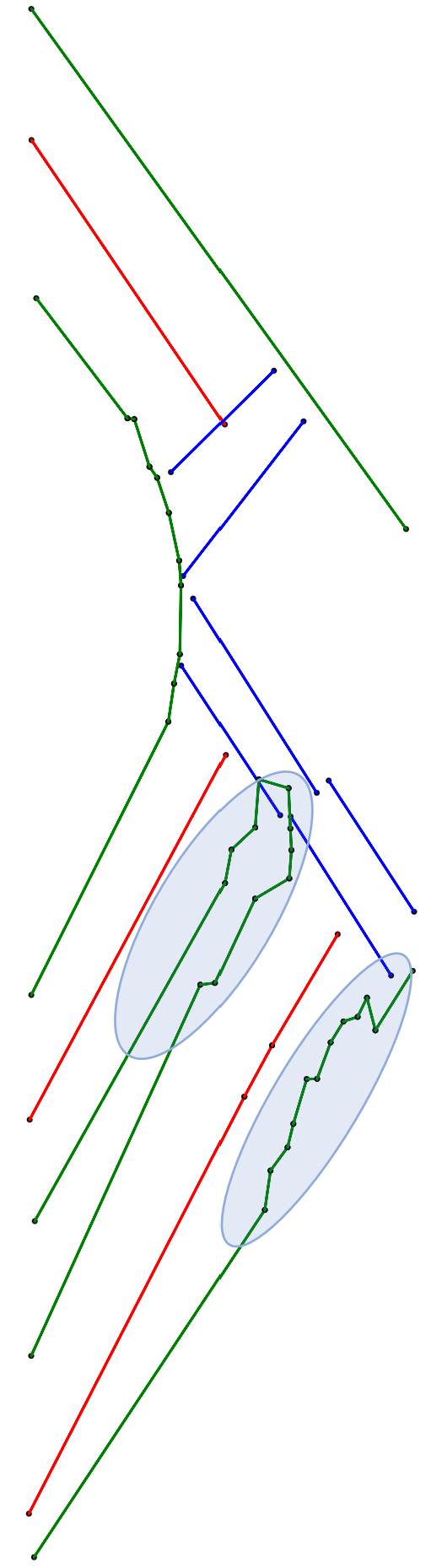}
       \includegraphics[width=0.12\textwidth, height = 160pt]{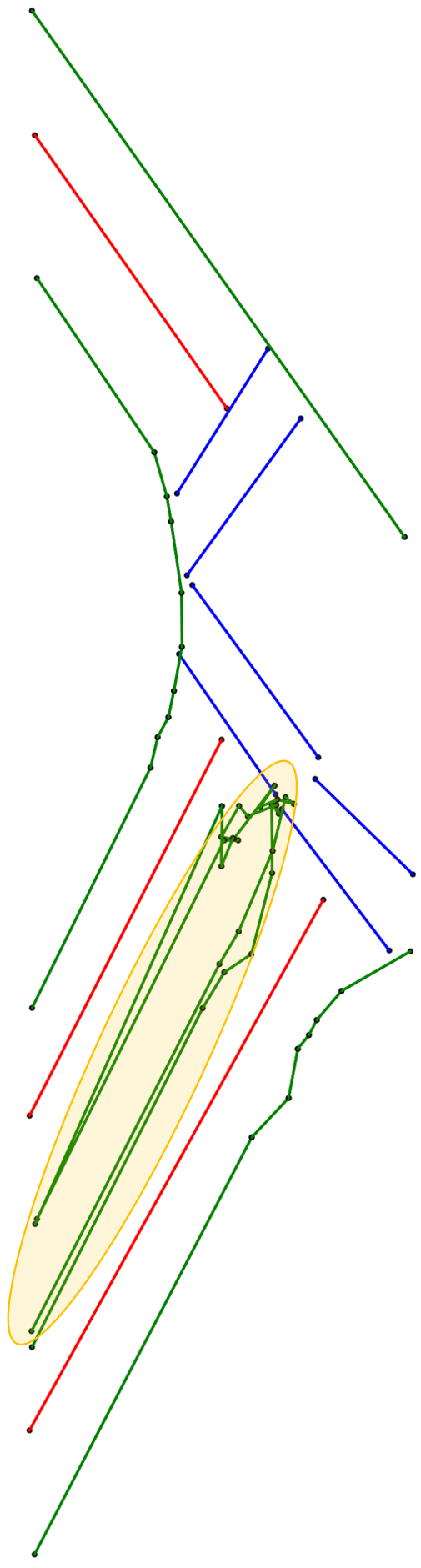}
       \includegraphics[width=0.12\textwidth, height = 160pt]{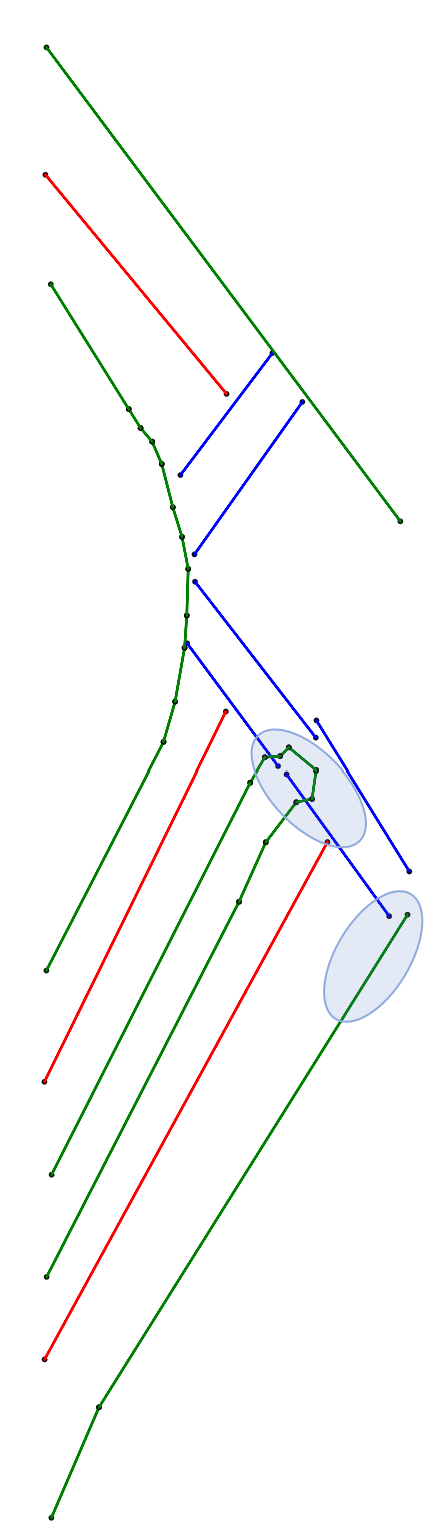}
       \includegraphics[width=0.12\textwidth, height = 160pt]{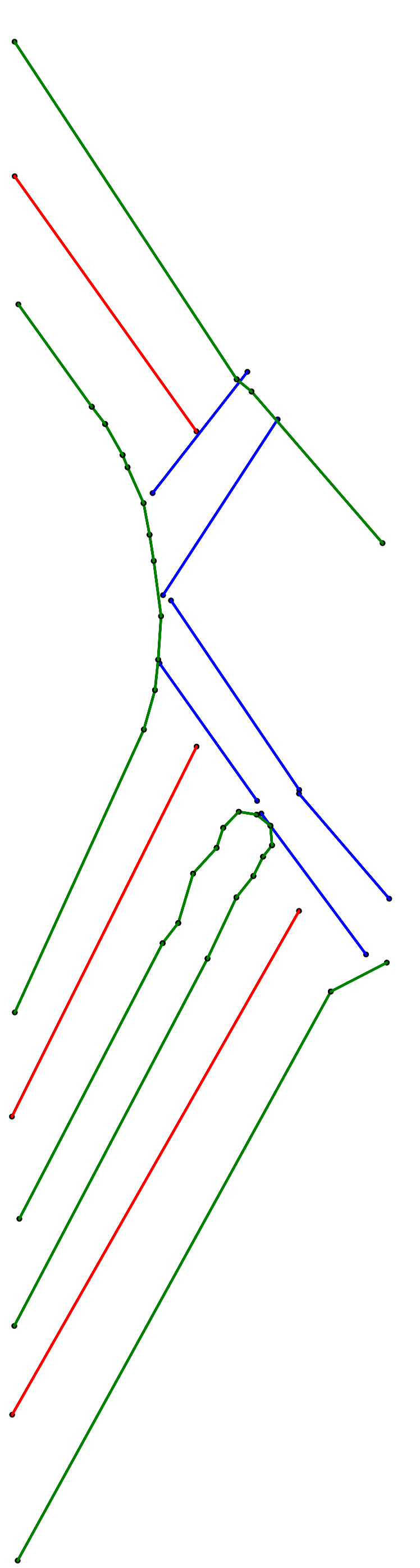}
        }
    \hfill
    \subfloat{
       \includegraphics[width=0.24\textwidth, height = 160pt]{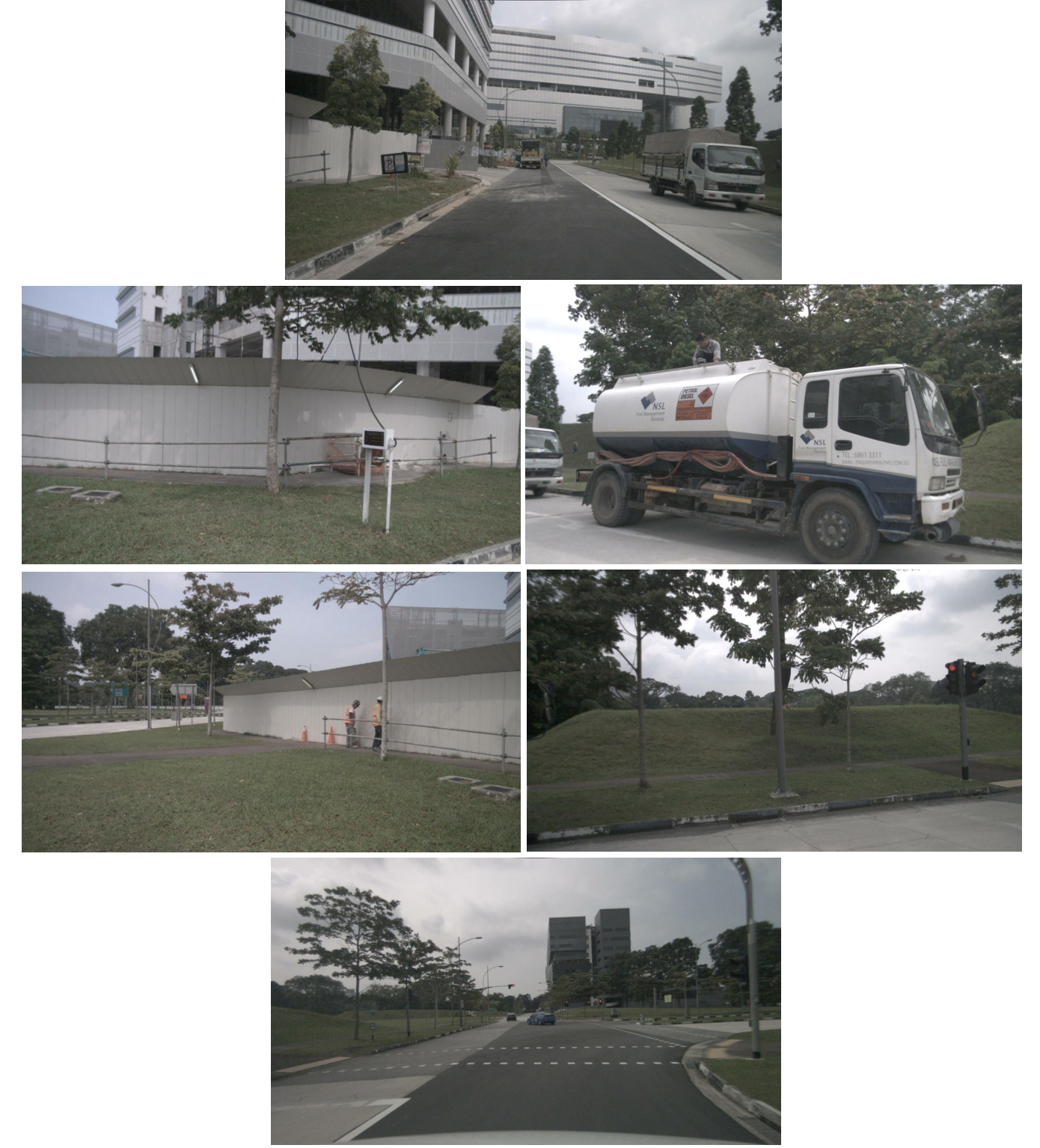}
       \includegraphics[width=0.12\textwidth, height = 160pt]{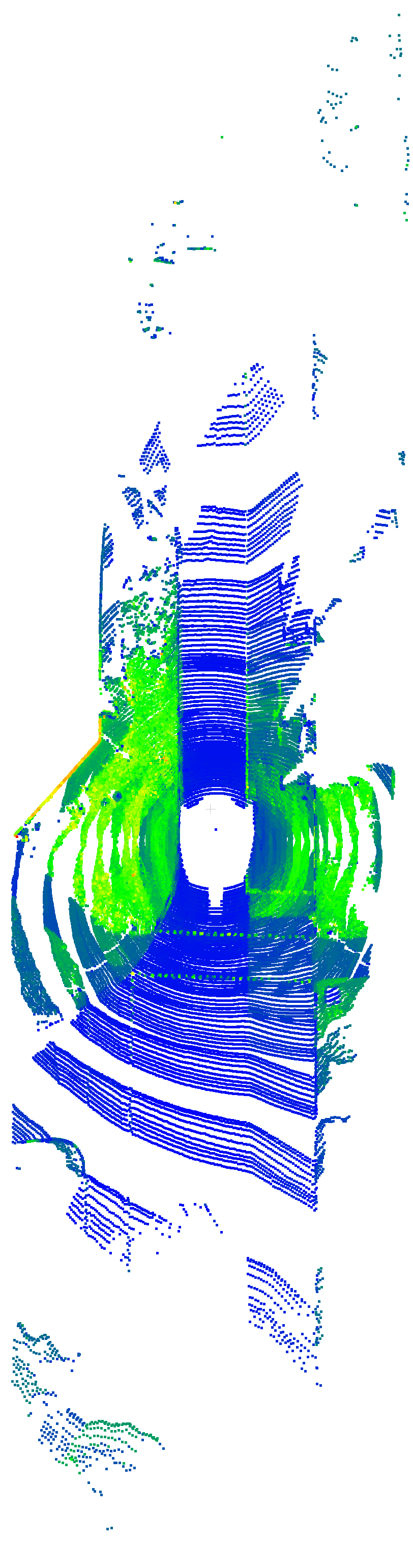}
       \includegraphics[width=0.12\textwidth, height =160pt]{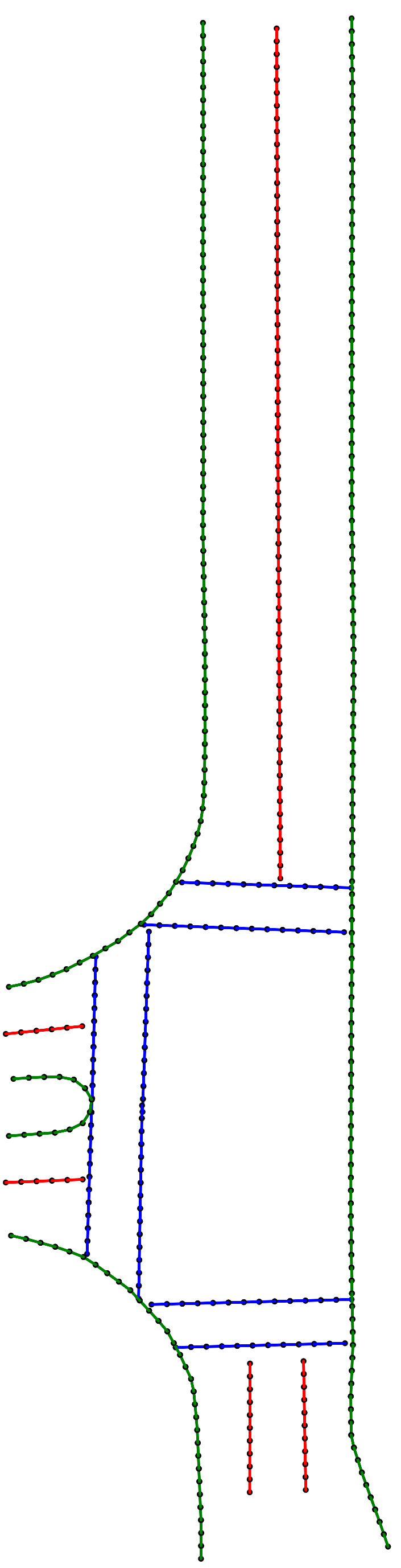}
       \includegraphics[width=0.12\textwidth, height = 160pt]{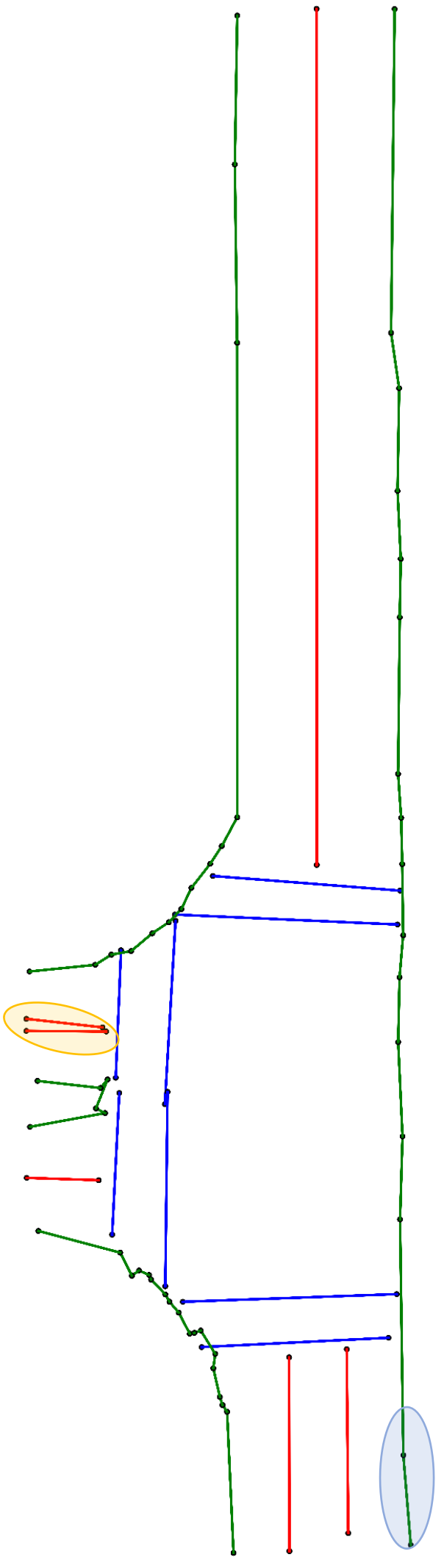}
       \includegraphics[width=0.12\textwidth, height = 160pt]{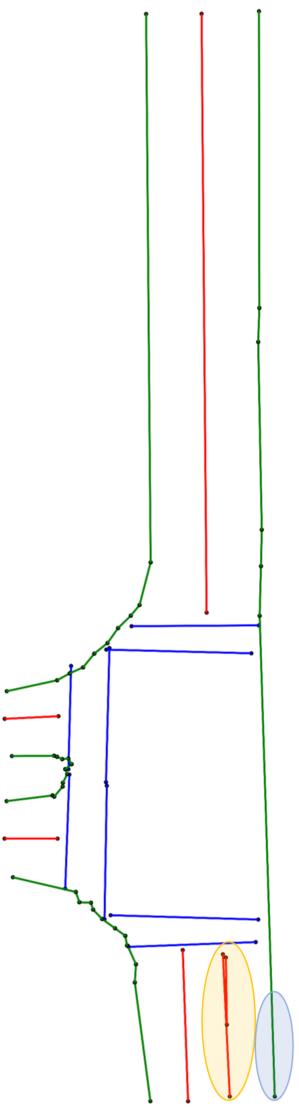}
       \includegraphics[width=0.12\textwidth, height = 160pt]{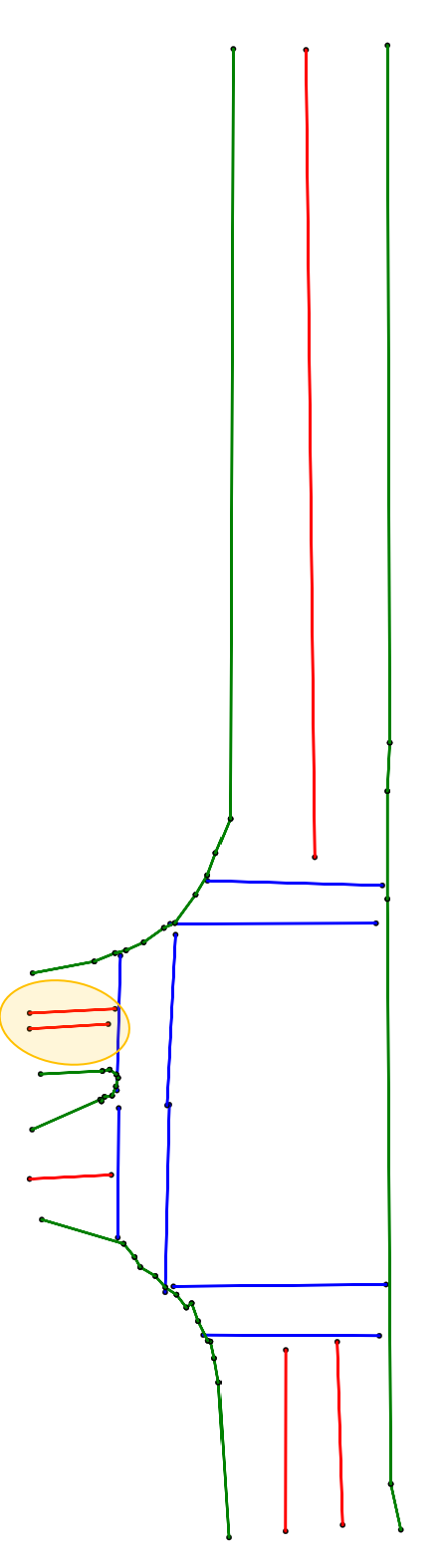}
       \includegraphics[width=0.12\textwidth, height = 160pt]{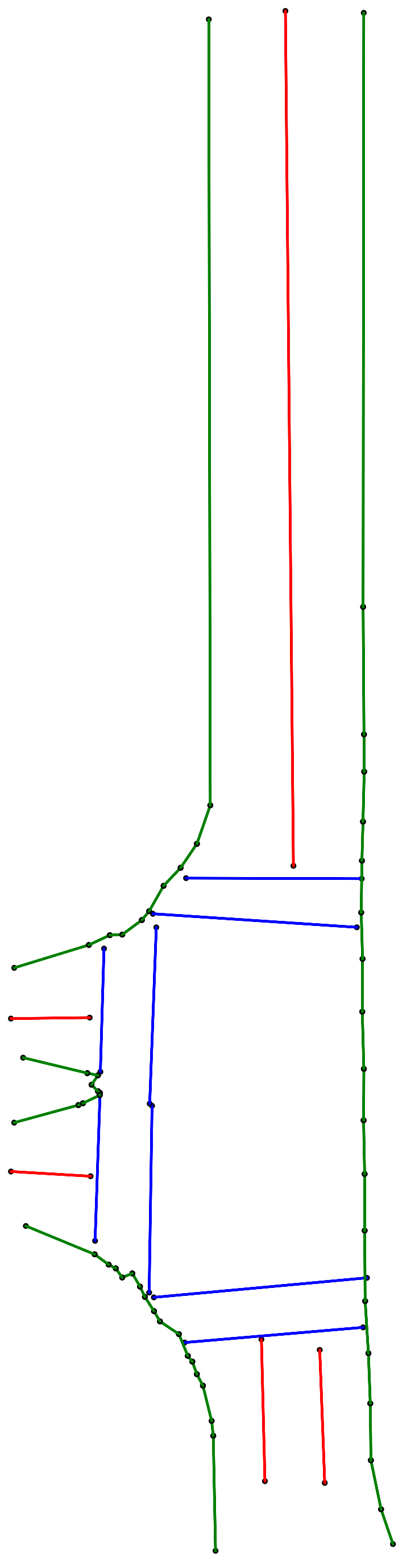}
        }
    \hfill
    \subfloat{
        \includegraphics[width=0.24\textwidth, height = 160pt]{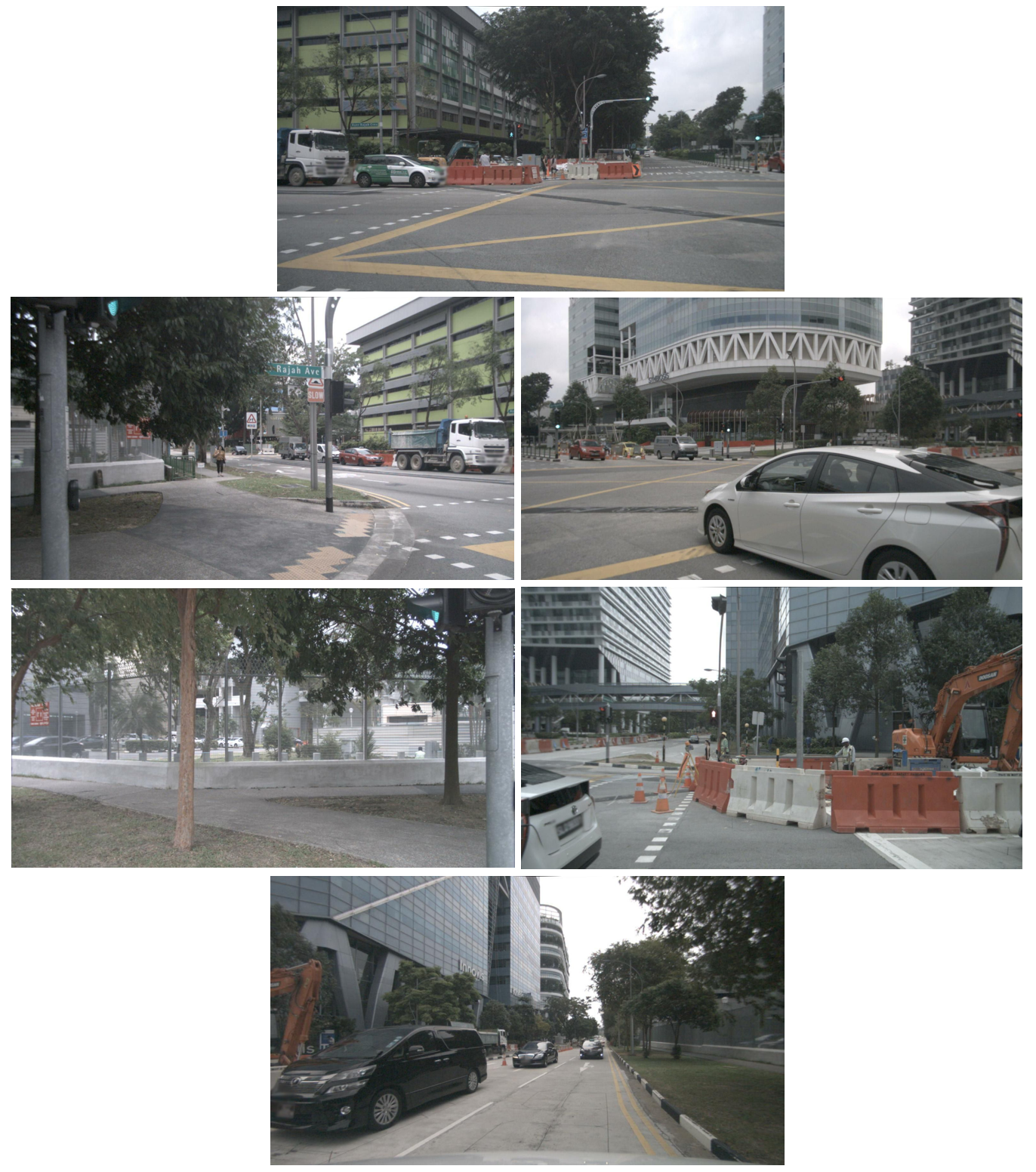}
        \includegraphics[width=0.12\textwidth, height = 160pt]{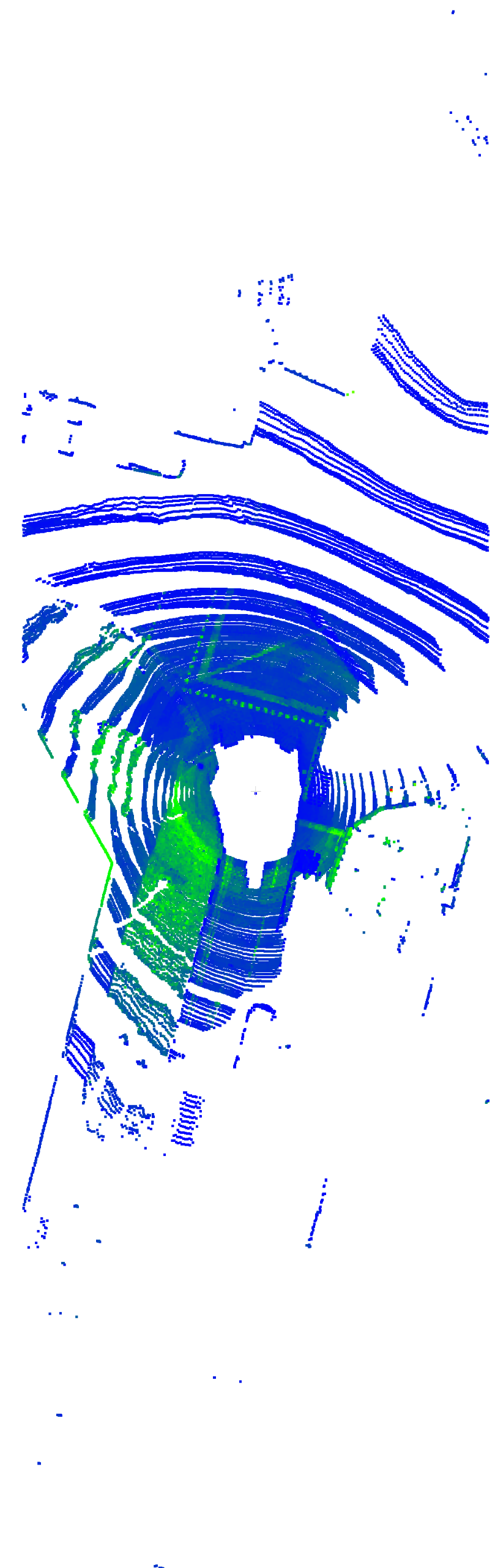}
        \includegraphics[width=0.12\textwidth, height =160pt]{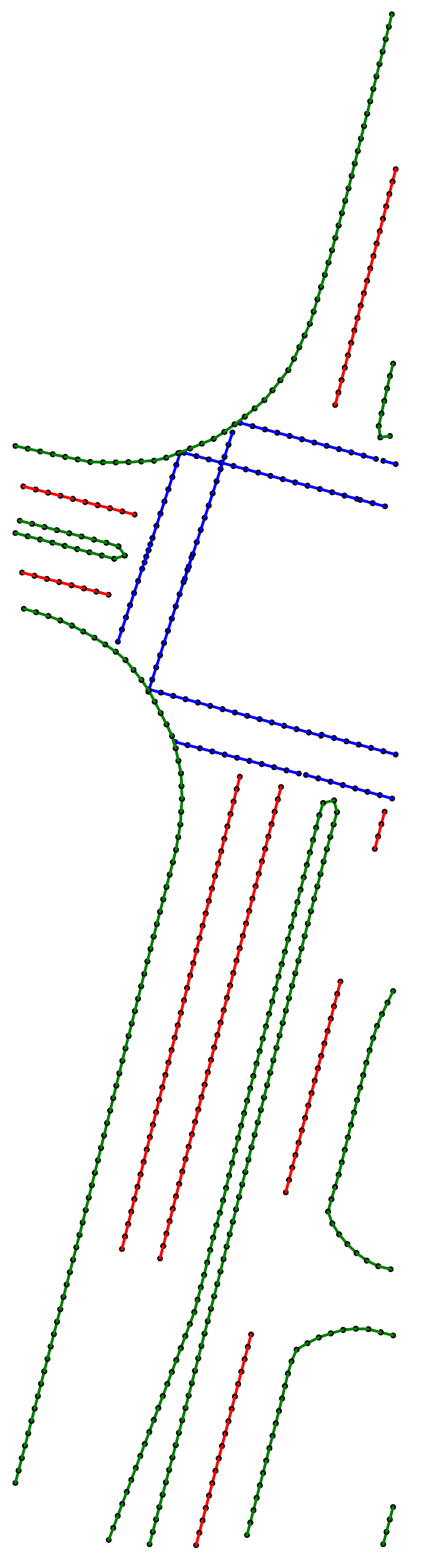}
        \includegraphics[width=0.12\textwidth, height = 160pt]{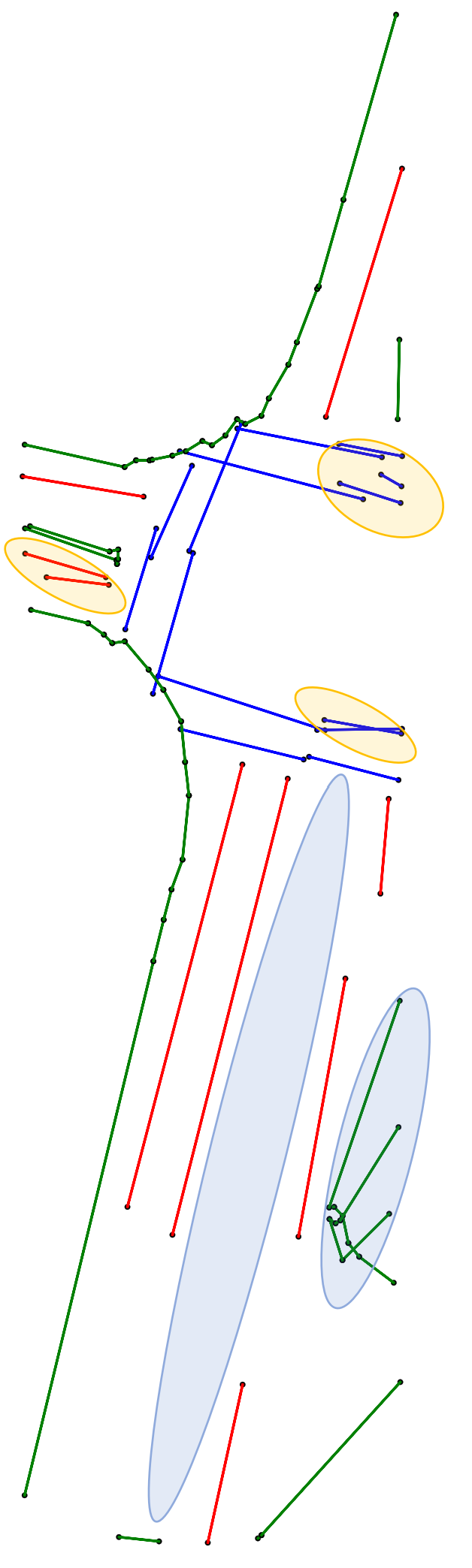} 
        \includegraphics[width=0.12\textwidth, height = 160pt]{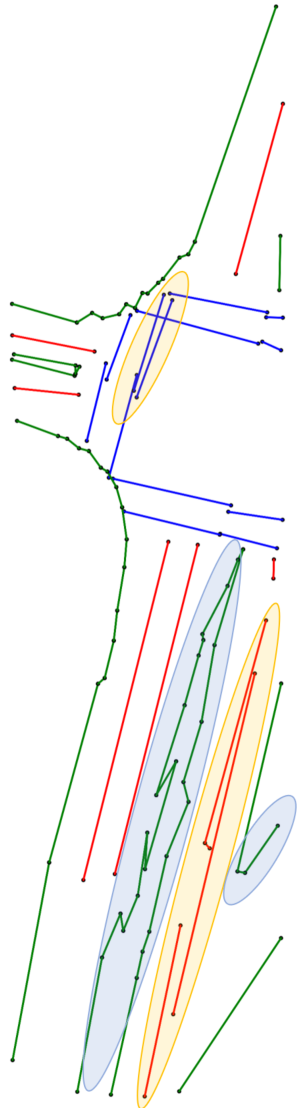}
        \includegraphics[width=0.12\textwidth, height = 160pt]{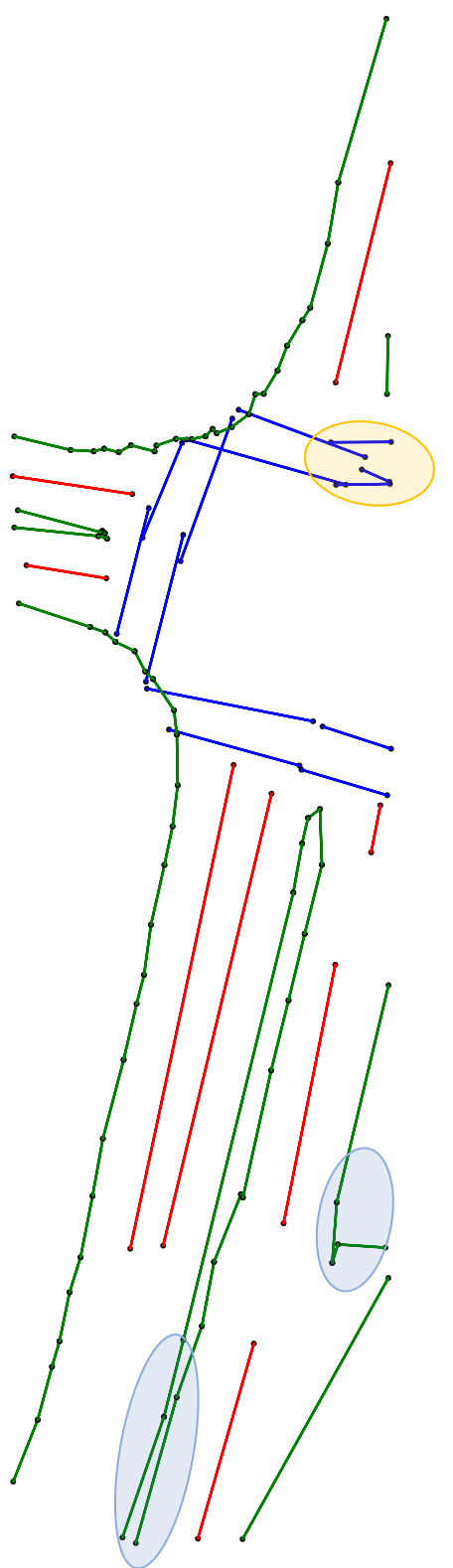}
        \includegraphics[width=0.12\textwidth, height = 160pt]{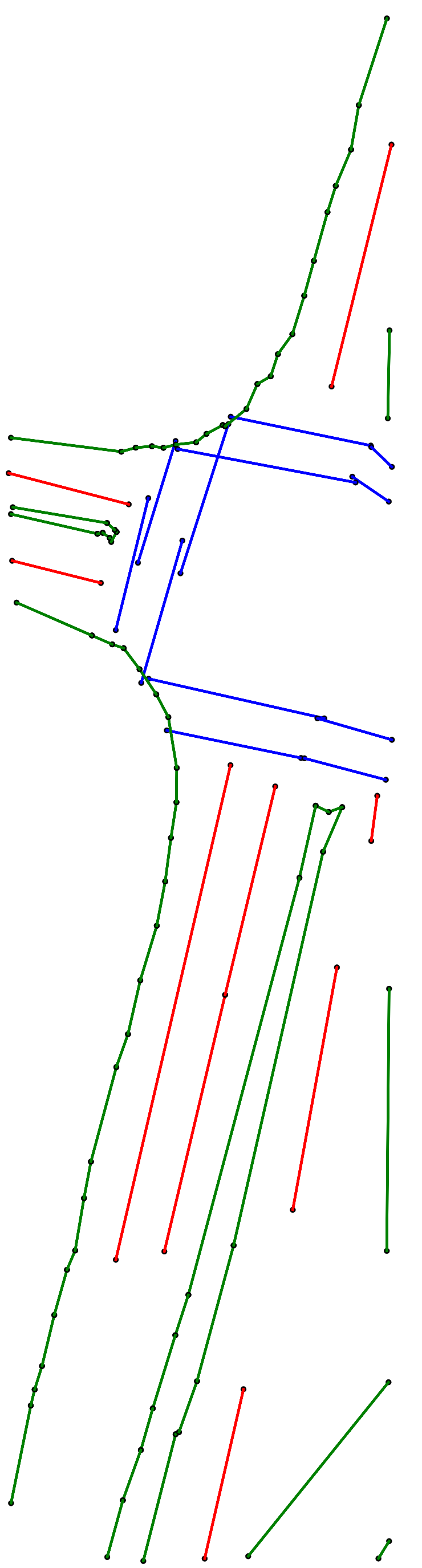}
        }
    \hfill \protect\\
    \caption{Visualization comparison between different modules of SuperMapNet on nuScenes dataset, where erroneous shapes are circled in blue and entanglement between elements are in yellow. Road boundaries are colored in green, while lane dividers and pedestrian crossings are in red and blue, respectively. Each contains six columns, (a) camera images; (b) LiDAR point clouds; (c) ground-truth; (d) baseline; (e) baseline only with SGC module; (f) baseline only with PEC module; and (g) baseline with both SGC and PEC modules.  }
    \label{fig:5}
\end{figure*}

Fig. \ref{fig:5} shows the visualization of our SuperMapNet on nuScenes dataset. It is obvious in Fig.(d) that, without SGC and PEC modules, there are erroneous shapes (e.g., road boundaries), entanglement between elements (e.g., pedestrian crossing), and element missing (e.g., road boundaries) in local maps predicted by baseline. By adding the proposed SGC module, element missing is disappeared in Fig. \ref{fig:5}(e), owing to the cross-attention based synergy enhancement which can fill the feature holes in the distance of LiDAR BEV features and enhance perception ability and ranges. However, erroneous shapes and entanglement between elements are still existed. Only using the PEC module can effectively handle the entanglement between elements as shown in Fig. \ref{fig:5}(f), as it incorporates three types of modeling information between points and elements. However, at the edge of the local map, erroneous shapes in details still persist due to the lack of perceptual capability enhanced by the SGC module. Fig. \ref{fig:5}(g) shows that, our SuperMapNet with both SGC module and PEC module can well address the problem of erroneous shapes, entanglement between elements and element missing, showing significant superiority.

\begin{figure*}[tp]
    \centering
    \subfloat{
        \includegraphics[width=0.36\textwidth, height = 160pt]{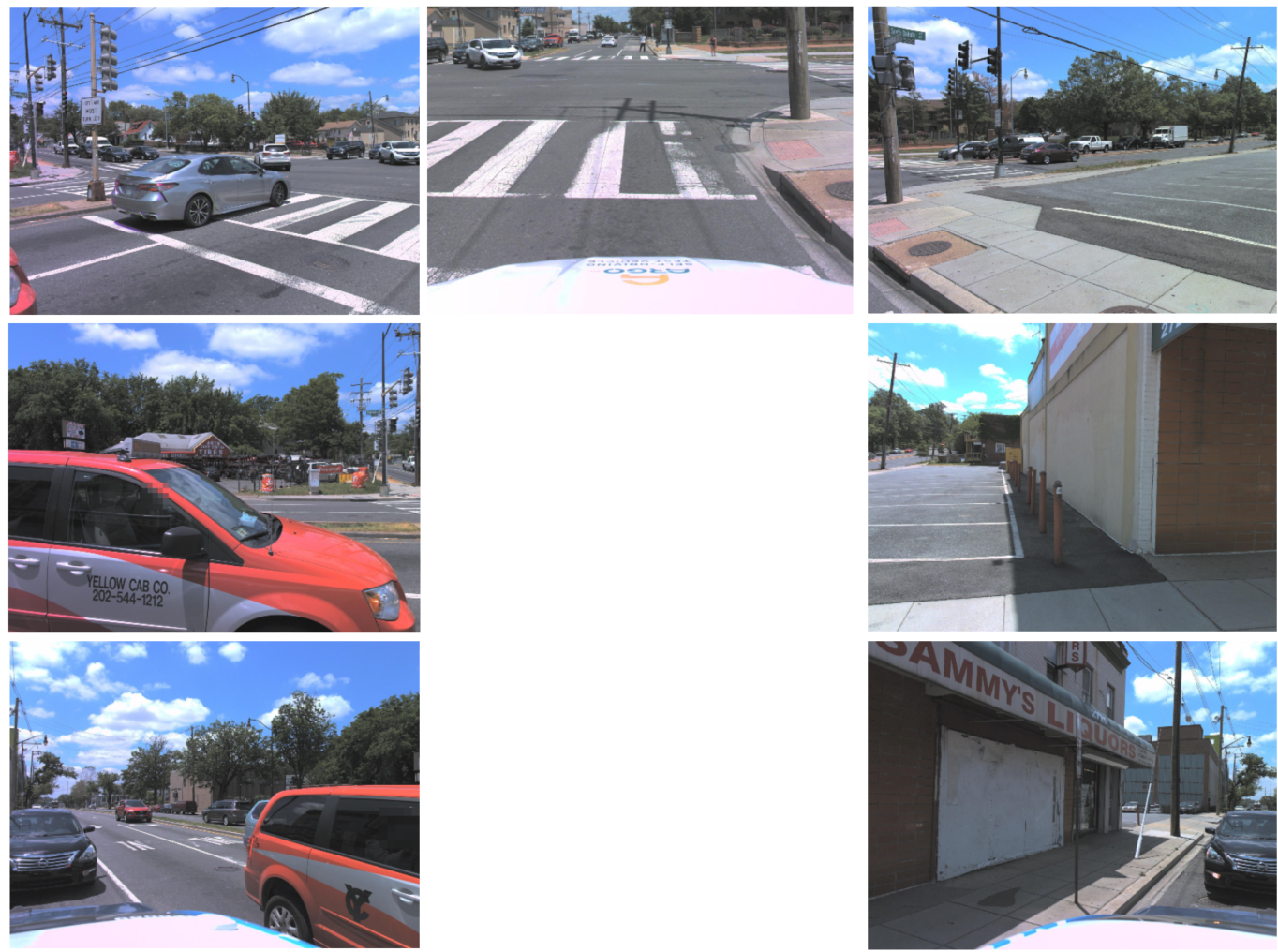}
   }
   \subfloat{
        \includegraphics[width=0.12\textwidth, height = 160pt]{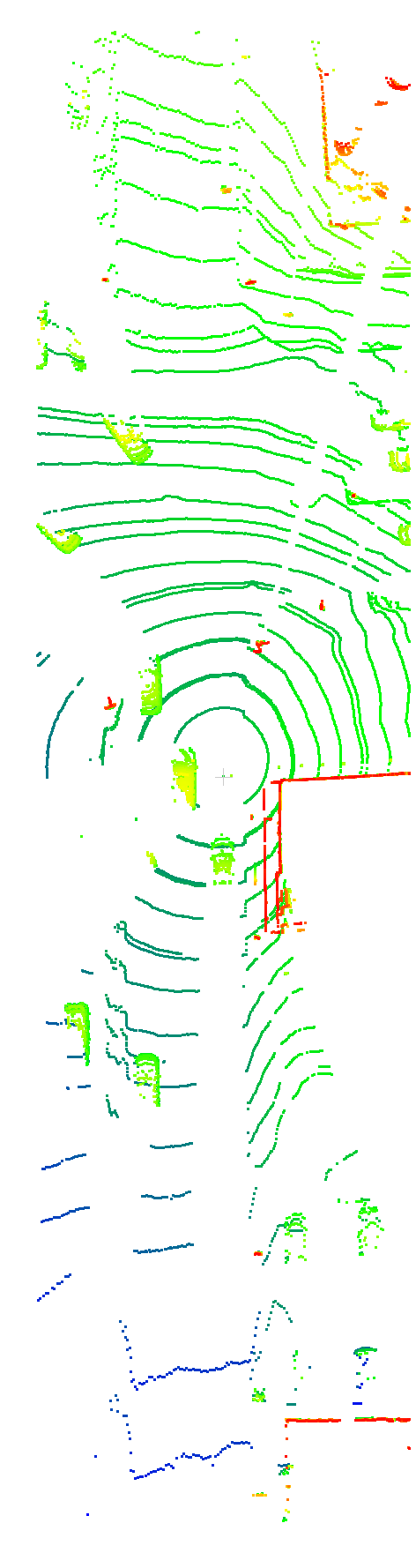}
   }
   \subfloat{
        \includegraphics[width=0.12\textwidth, height=160pt ]{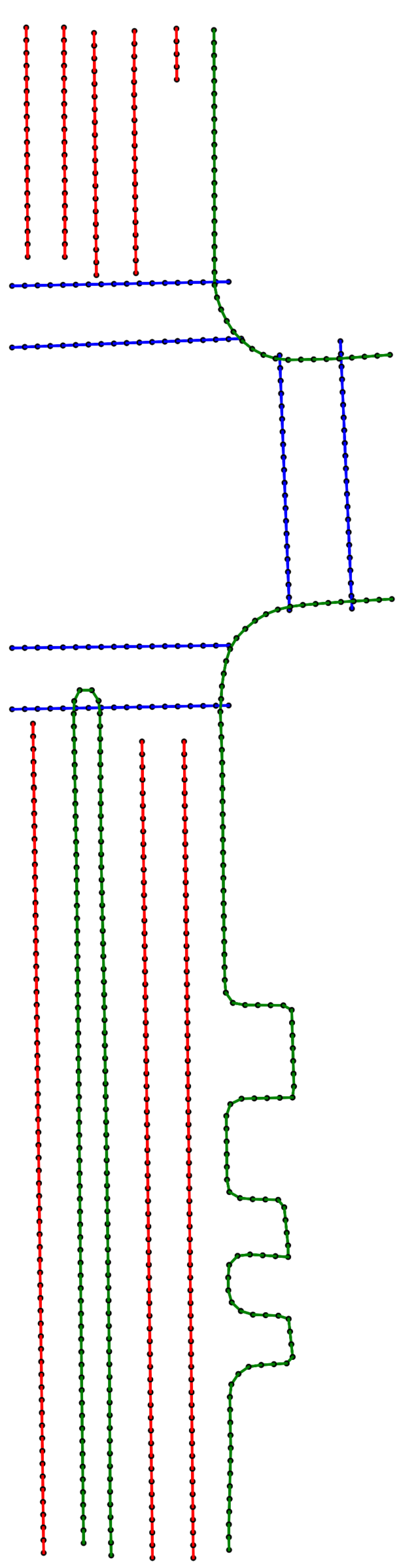}
   }
   \subfloat{
         \includegraphics[width=0.12\textwidth ,height=160pt]{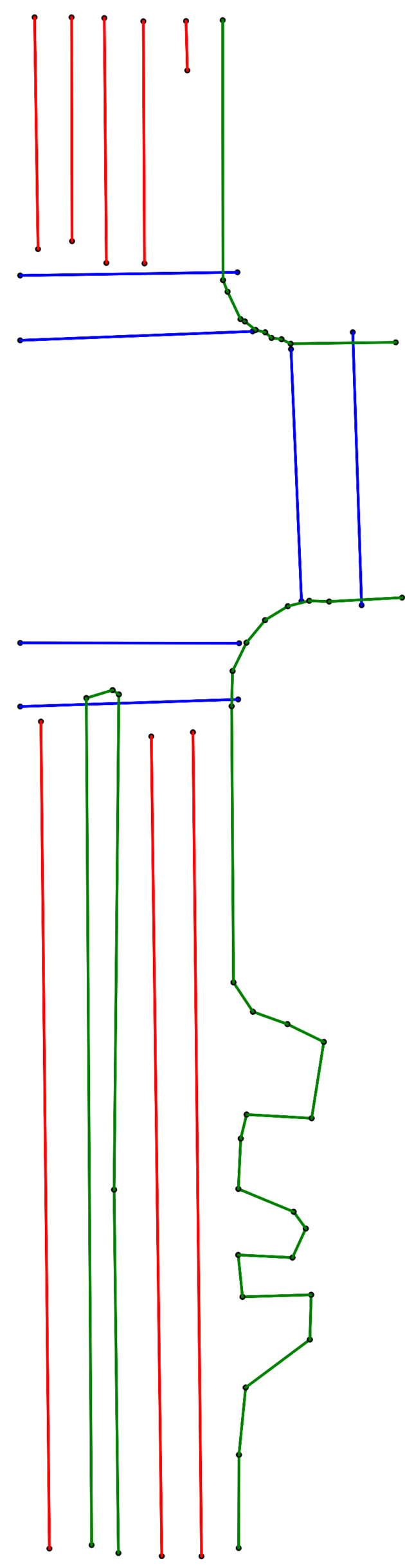}
   }
   \hfill
   \subfloat{
        \includegraphics[width=0.36\textwidth, height = 160pt]{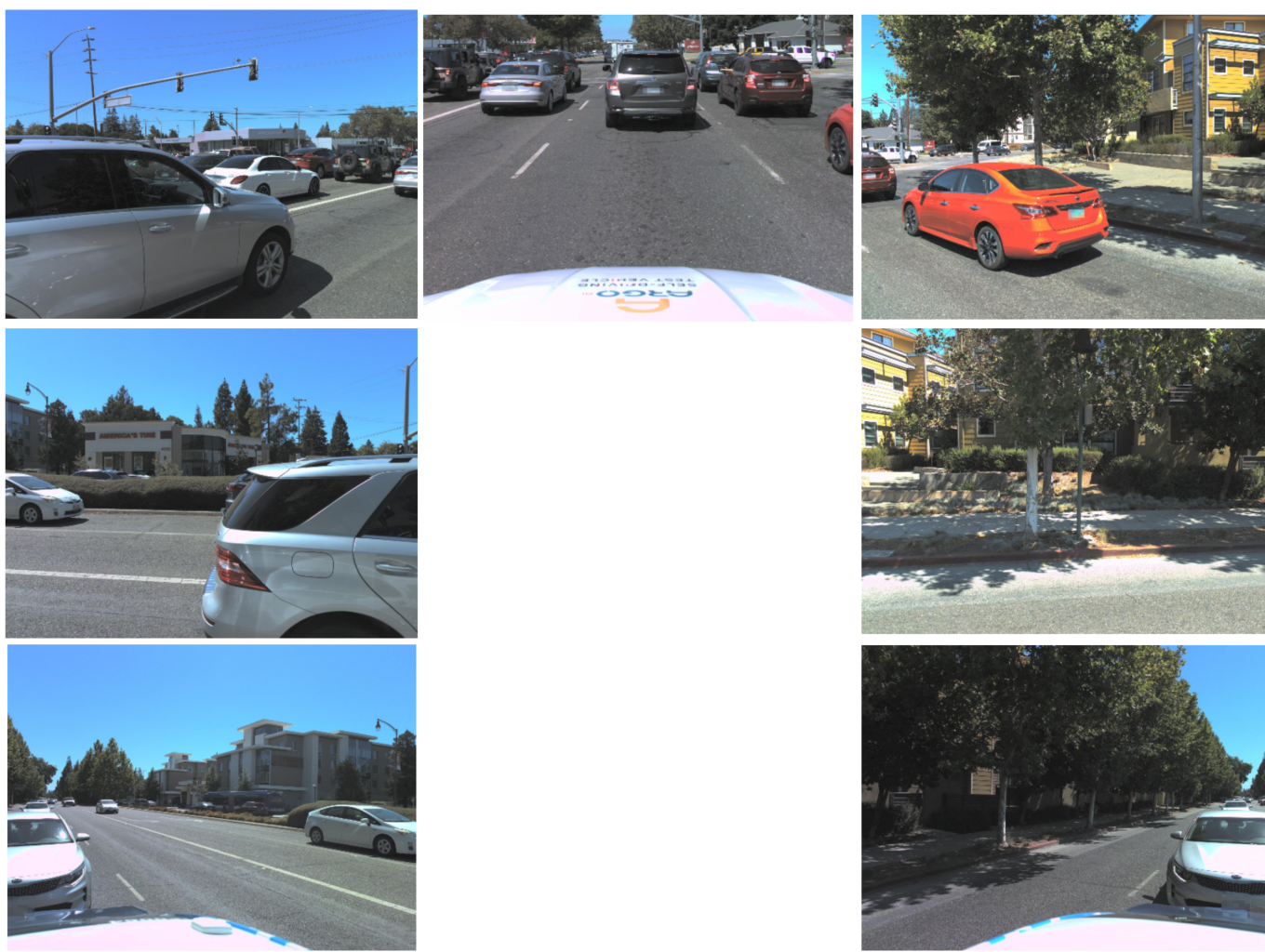}
   }
   \subfloat{
        \includegraphics[width=0.12\textwidth, height = 160pt]{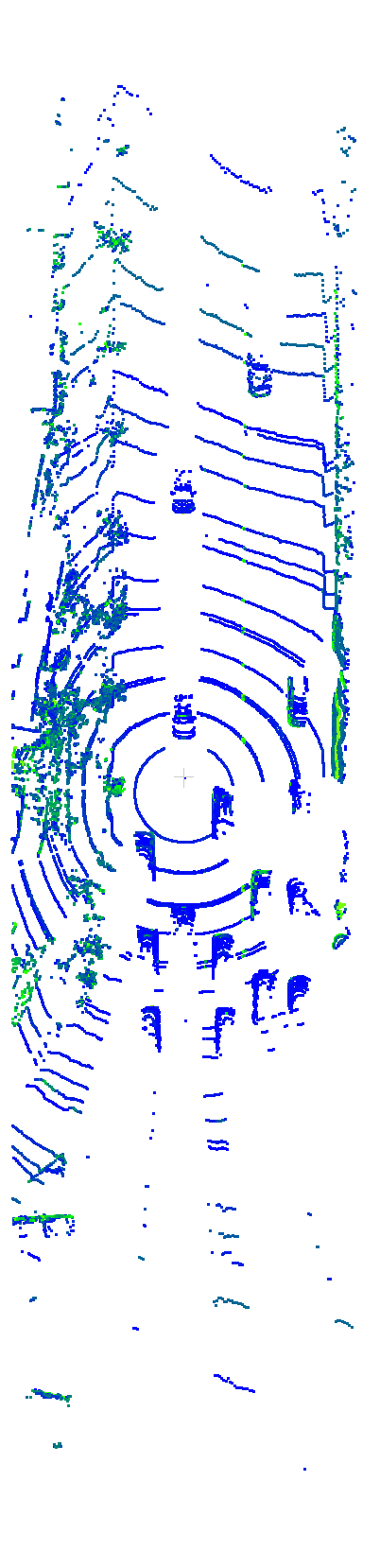}
   }
   \subfloat{
        \includegraphics[width=0.12\textwidth, height=160pt ]{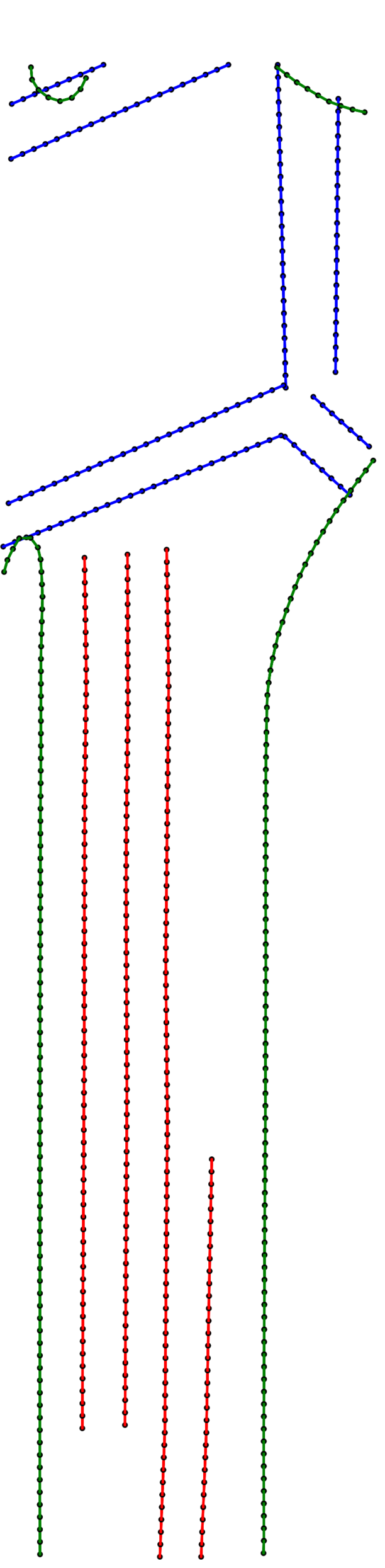}
   }
   \subfloat{
         \includegraphics[width=0.12\textwidth ,height=160pt]{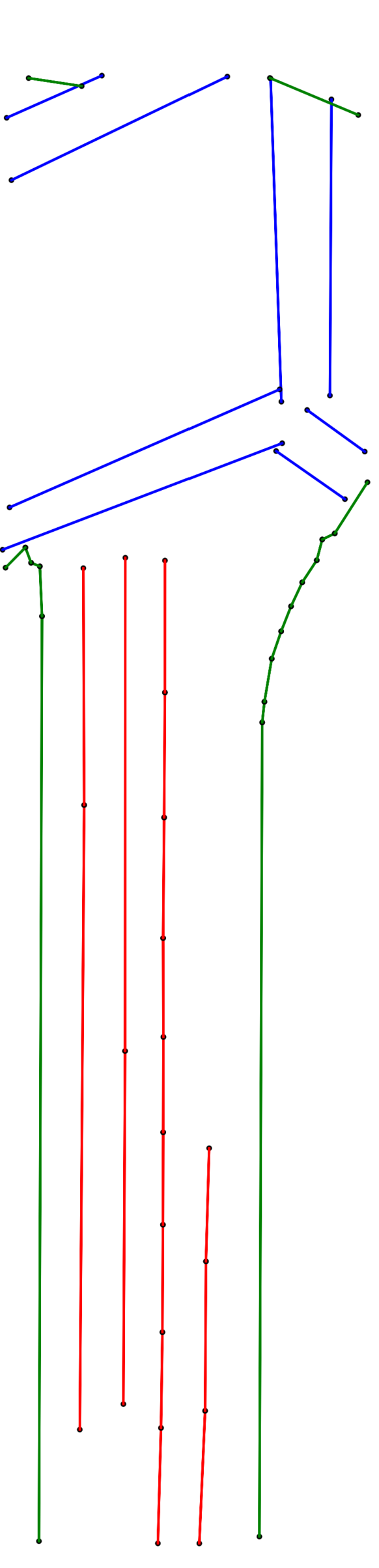}
   }
    \hfill
   \subfloat{
        \includegraphics[width=0.36\textwidth, height = 160pt]{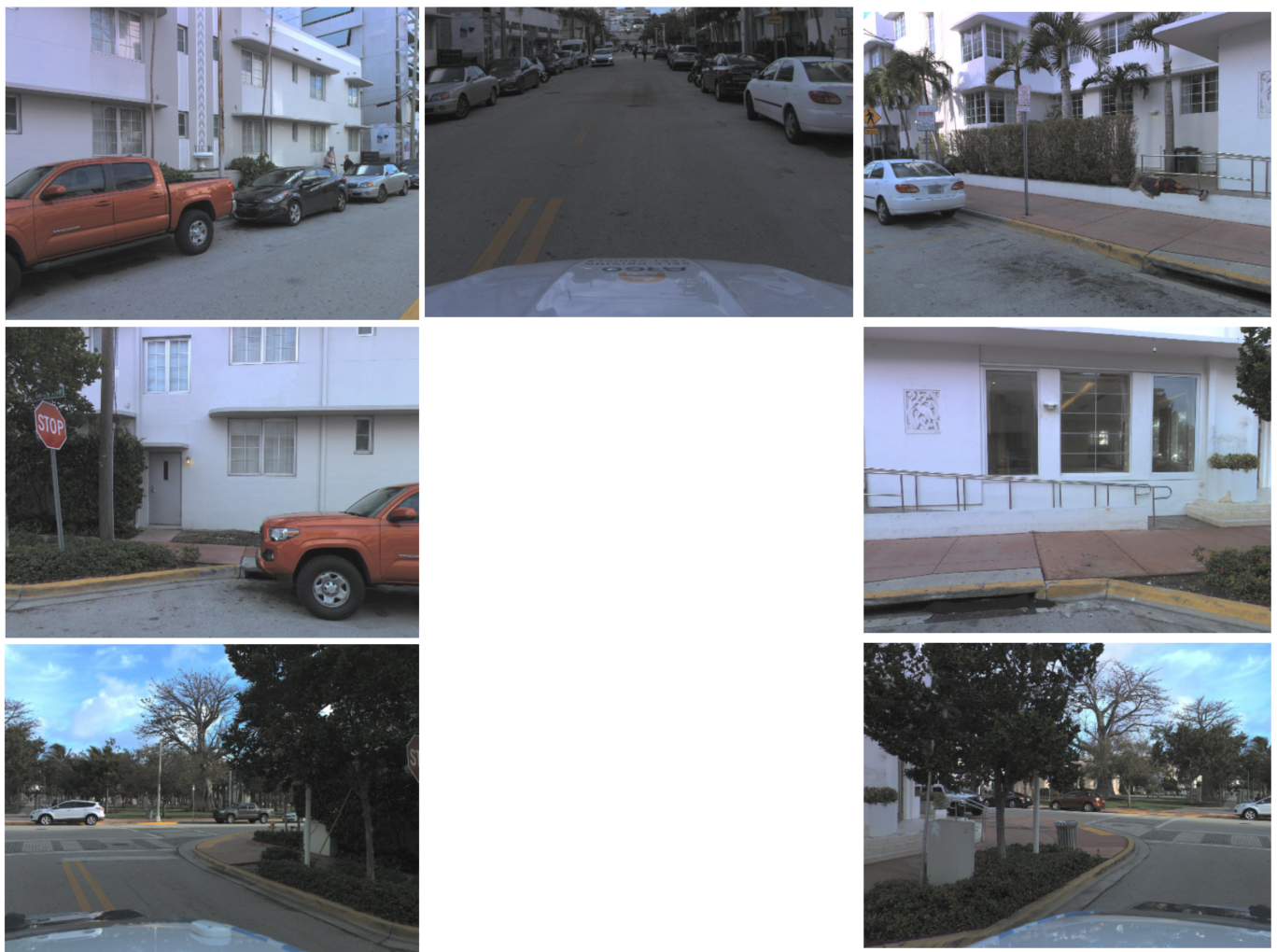}
   }
   \subfloat{
        \includegraphics[width=0.12\textwidth, height = 160pt]{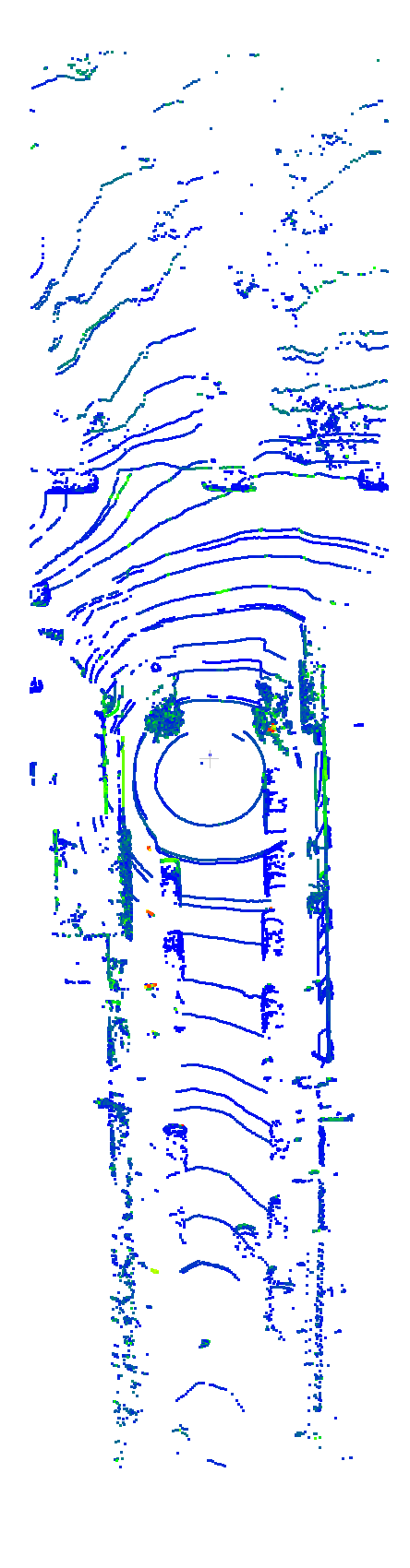}
   }
   \subfloat{
        \includegraphics[width=0.12\textwidth, height=160pt ]{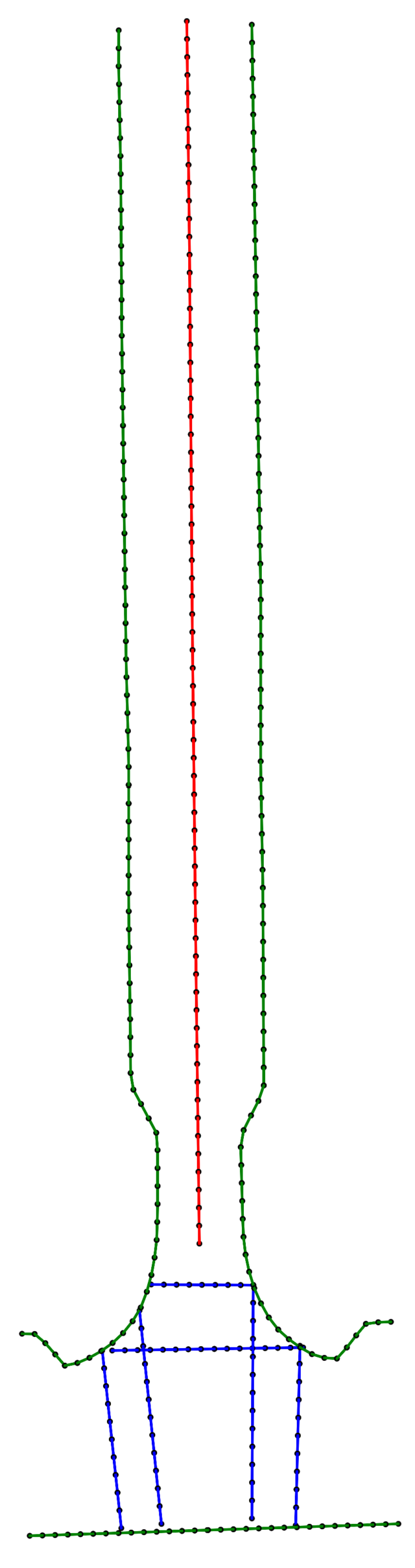}
   }
   \subfloat{
         \includegraphics[width=0.12\textwidth ,height=160pt]{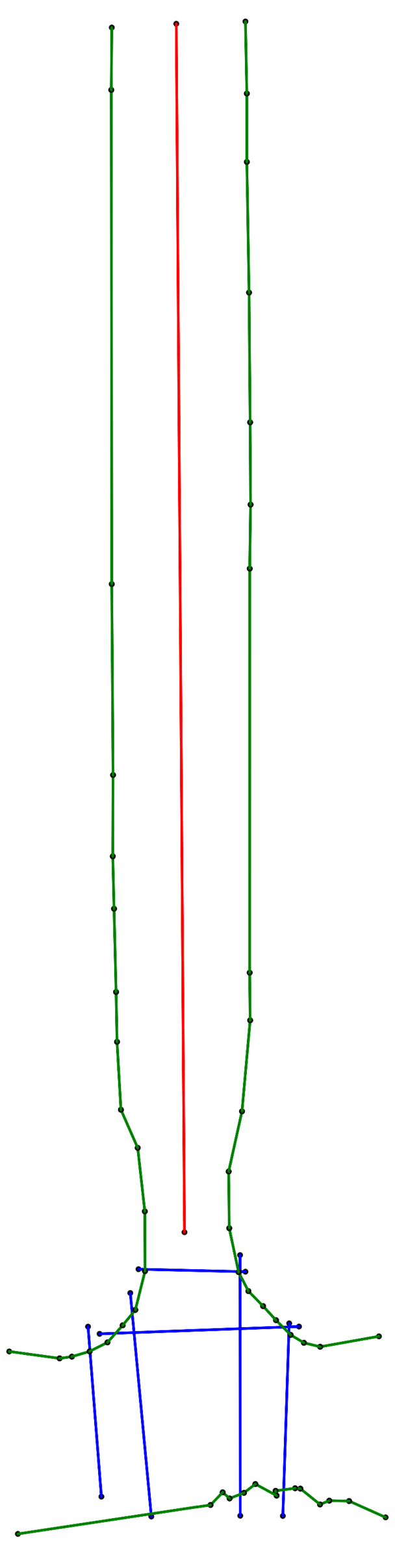}
   }
    \caption{Visualization of SuperMapNet on Argoverse2 dataset. Road boundaries are colored in green, while lane dividers and pedestrian crossings are in red and blue, respectively. Each contains four columns, (a) camera images; (b) LiDAR point clouds; (c) ground-truth; (d) results predicted by SuperMapNet with both SGC and PEC modules. }
    \label{fig:arg2}
\end{figure*}

Fig. \ref{fig:arg2} shows the visualization of our SuperMapNet on Argoverse2 dataset. As shown in Fig. \ref{fig:arg2}(a) and Fig. \ref{fig:arg2}(b), each frame on Argoverse2 dataset contains from 7 ring cameras, and the LiDAR point cloud is much sparser than nuScenes dataset, resulting in point clouds being unable to provide more spatial information in details. In addition, the phenomenon of occlusions caused by other vehicles in the dataset is more severe, resulting in feature holes in BEV sapce. Fig. \ref{fig:arg2}(d) show that, although boundaries in Argoverse2 dataset are more irregular and the distribution and shapes of pedestrian crossings are more complex, our SuperMapNet can still well construct all map elements and well handle erroneous element shapes or entanglement between elements in most cases.

\section{Conclusion}
\label{sec:conclusion}

SuperMapNet is an effective network designed for long-range and high-accuracy vectorized HD map construction. It boasts a robust coupling between semantic and geometric information with the consideration of synergies and disparities, as well as a coupling between point and element information at three levels, Point2Point, Element2Element, and Point2Element. Extensive experiments have demonstrated the significant potential of our SuperMapNet, setting new SOTAs on both nuScenes and Argoverse2 datasets. We are confident that SuperMapNet offers a novel perspective for future research in the vectorized HD map construction tasks.

\appendix

\section*{CRediT authorship contribution statement}
Ruqin Zhou: Conceptualization, Methodology, Software, Writing – original draft, Writing–review \& editing. Chenguang Dai: Funding acquisition, Project administration, Resources. Wanshou Jiang: Conceptualization, Investigation, Resources. Yongsheng Zhang: Funding acquisition, Resources. Hanyun Wang: Validation, Writing–review \& editing. San Jiang: Conceptualization, Software, Validation, Writing–review \& editing, Funding acquisition.

\section*{Declaration of competing interest}
The authors declare that they have no known competing financial interests or personal relationships that could have appeared to influence the work reported in this paper.

\section*{Acknowledgments}
This research was funded by the National Natural Science Foundation of China (Grant No. 42371442), the Henan Provincial Natural Science Foundation of China (Grant No. 252300420876).
\bibliographystyle{elsarticle-harv} 
\bibliography{cas-refs}





\end{document}